\documentclass[10pt,twocolumn]{article}
\usepackage{arxivshim}   
\usepackage{cite}
\usepackage{amsmath,amssymb,amsfonts}
\usepackage{algorithmic}
\usepackage{graphicx}
\usepackage{textcomp}

\usepackage{microtype}

\usepackage{subcaption}
\usepackage{pifont}

\usepackage{booktabs} 
\usepackage{tabularx}
\usepackage{array}
\usepackage{multirow}
\usepackage{bm}
\usepackage{hyperref}

\makeatletter
\AtBeginDocument{\DeclareMathVersion{bold}
\SetSymbolFont{operators}{bold}{T1}{times}{b}{n}
\SetSymbolFont{NewLetters}{bold}{T1}{times}{b}{it}
\SetMathAlphabet{\mathrm}{bold}{T1}{times}{b}{n}
\SetMathAlphabet{\mathit}{bold}{T1}{times}{b}{it}
\SetMathAlphabet{\mathbf}{bold}{T1}{times}{b}{n}
\SetMathAlphabet{\mathtt}{bold}{OT1}{pcr}{b}{n}
\SetSymbolFont{symbols}{bold}{OMS}{cmsy}{b}{n}
\renewcommand\boldmath{\@nomath\boldmath\mathversion{bold}}}
\makeatother

\def\BibTeX{{\rm B\kern-.05em{\sc i\kern-.025em b}\kern-.08em
    T\kern-.1667em\lower.7ex\hbox{E}\kern-.125emX}}

\begin{document}
\doi{10.1109/ACCESS.xxxx.xxxxxxx}

\title{STAG-Net: Structure-Aware Attention over Landmark Groups for Lightweight Head Pose Estimation}
\author{\uppercase{Asiri Gawesha Lindamulage}\authorrefmark{1},
\uppercase{Sanka Mohottala}\authorrefmark{2}, \uppercase{Dharshana Kasthurirathna}\authorrefmark{3}\IEEEmembership{Member, IEEE}, AND \uppercase{Nuwan Kodagoda}\authorrefmark{4}}

\address[1]{Faculty of Engineering, Sri Lanka Institute of Information Technology, Sri Lanka (e-mail: asiri.l@sliit.lk)}
\address[2]{Faculty of Engineering, University of Sri Jayewardenepura, Sri Lanka (e-mail: sanka@sjp.ac.lk)}
\address[3]{University of Colombo School of Computing, Colombo 07, Sri Lanka (e-mail: dka@ucsc.cmb.ac.lk)}
\address[4]{Faculty of Computing, Sri Lanka Institute of Information Technology, Sri Lanka (e-mail: nuwan.k@sliit.lk)}

\tfootnote{This work was supported by the Accelerating Higher Education Expansion
and Development (AHEAD) Operation of the Ministry of
Higher Education of Sri Lanka, funded by the World Bank
(\url{https://ahead.lk/result-area-3/}) and  a research grant funded by the Sri Lanka Institute of Information Technology, Sri Lanka (Grant No.~PVC(R\&I)/RG/2025/12).}

\markboth
{Lindamulage \headeretal: STAG-Net: Structure-Aware Attention over Landmark Groups for Lightweight Head Pose Estimation Pose Estimation}
{Lindamulage \headeretal: STAG-Net: Structure-Aware Attention over Landmark Groups for Lightweight Head Pose Estimation}

\corresp{Corresponding author: Asiri Gawesha Lindamulage (e-mail: asiri.l@sliit.lk).}

\begin{abstract}
Head pose estimation is important for human computer interaction and driver monitoring, but many image based methods remain too costly for resource constrained deployment. We propose STAG-Net, a structure-aware attention over landmark groups for lightweight head Pose estimation that estimates head pose from facial landmarks using movement aware selection, spatial clustering, compact DeepSet encoders, self attention, and a global geometry descriptor. We develop variants using FaceMesh and Face Alignment Network landmarks and evaluate them under cross dataset and subject independent protocols on 300W-LP, AFLW2000, and BIWI. STAG-Net achieves the lowest mean absolute error among all compared landmark based methods in every evaluated setting. On BIWI frames with valid detections, the FAN variant also records the lowest overall error among all compared methods, reaching $3.51^\circ$ under cross dataset evaluation and $2.39^\circ$ under subject independent evaluation. The regressors use only 0.012 to 0.043 million parameters and 24.1 to 83.6 thousand FLOPs, about three orders of magnitude fewer FLOPs than FSA-Net. Despite including landmark extraction and regression, the FaceMesh pipeline reaches 322.9 frames per second on a Tesla T4 GPU, nearly twice the 164.5 frames per second of the fastest measured image based baseline, and 111.0 frames per second on CPU. After normalization by dataset size, STAG-Net processes training samples seven to eight times faster than FSA-Net and more than one hundred times faster than HopeNet and TokenHPE, while complete training finishes within 28 minutes. These results show that STAG-Net combines leading landmark based accuracy with substantially lower training and inference cost.

\end{abstract}


\titlepgskip=-21pt

\maketitle

\section{Introduction}
\label{sec:introduction}

\subsection{Background}

Head pose estimation (HPE) predicts the orientation of a human head, usually in terms of yaw, pitch, and roll, from an image or a geometric facial representation. It supports applications such as human computer interaction, driver monitoring, augmented and virtual reality, and behavioral analysis~\cite{a1,Fine-Grained}. Many accurate HPE systems rely on convolutional or transformer based processing of high resolution facial images. This dense visual processing increases memory use and computational cost, which limits deployment on low power edge devices.

Facial landmarks offer a more compact alternative because they reduce the input to a small set of geometric coordinates while preserving the facial structure needed for pose estimation~\cite{howfar}. A compact landmark based regressor can also support local and privacy preserving inference without sending images to a cloud service. These properties are attractive for wearable systems, IoT devices, and TinyML applications~\cite{warden2019tinyml}. However, the complete deployment cost also depends on the landmark detector, so both the regression model and the combined pipeline must be evaluated.

Existing landmark based methods do not fully resolve this challenge. Many represent landmarks as a fixed order coordinate vector and apply a simple regressor, which limits explicit modeling of structural relationships~\cite{Wer17,Der17}. Graph based methods model interactions among facial points more directly~\cite{eva,osgnet,singlegnn}, but lightweight landmark native architectures that capture both local relationships within facial regions and global relationships across regions remain limited~\cite{ds_hpe,gads}. This motivates a compact structure aware model that retains the efficiency of landmark inputs while improving relational modeling and reporting both regressor level and combined pipeline costs.

To address these limitations, we propose STAG-Net (Structure-Aware Attention over Landmark Groups for Lightweight Head Pose Estimation), a landmark based lightweight
architecture for head pose estimation. STAG-Net partitions facial
landmarks into spatially coherent clusters discovered automatically
through connectivity constrained clustering, and applies a
movement aware ranking strategy that retains only the most
pose informative landmarks from each region. Each cluster is encoded
as an unordered set by a compact DeepSet module, and a multihead
attention mechanism models dependencies across clusters, allowing the
network to capture both local and global structural information. In
this way, STAG-Net overcomes the limitations of conventional
landmark based methods that rely on fixed order vector
representations or simple global aggregation.

Efficiency is treated as a primary design objective. The pose regressor uses permutation invariant set encoding and compact architectural components to balance prediction accuracy with model size and computation. We therefore evaluate not only the regressor, but also the landmark detector and regressor pipeline. This distinction is important because the compact regression stage is compatible with microcontroller class resource budgets, whereas the feasibility of a fully embedded system still depends on the upstream landmark detector and on device implementation.

The main contributions of this work are summarized as follows:

\begin{itemize}
    \item We propose STAG-Net, a compact landmark based head pose
    estimation architecture that combines automatically discovered
    landmark clusters, permutation invariant encoding within each
    cluster, attention across clusters, and a global geometry
    descriptor.
    \item We introduce a movement aware grouping and selection strategy that retains pose sensitive landmarks from every facial region. Through a design analysis, we examine the effects of the retained landmark percentage, connectivity graph construction, architectural hyperparameters, and learning rate scheduling strategies on estimation accuracy.
    \item Across the three reported evaluation settings, STAG-Net achieves the lowest reported MAE among the compared landmark based methods. On BIWI frames with valid landmark detections, it also obtains the lowest reported overall MAE among all compared methods under both protocols.  We further evaluate model size, regression FLOPs, landmark and regression pipeline throughput, and training cost, while using Pareto analysis to study the joint relationship between accuracy and model size.
    
    \item We analyze the operating ranges of the two landmark
    detectors and show that extreme pose coverage is mainly limited
    by landmark detection rather than pose regression. We also
    identify 36 physically implausible AFLW2000 annotations and report
    the cleaned angle ranges used in our experiments.
\end{itemize}

The remainder of this paper follows the same progression. Section~\ref{sec:related_work} reviews image based, landmark based, and efficient head pose estimation. Section~\ref{sec:method} presents the complete STAG-Net method, from landmark preprocessing and movement aware selection to grouped set encoding, attention, geometry fusion, and pose regression. Section~\ref{sec:experimental_setup} defines the datasets, protocols, metrics, and implementation details. Section~\ref{sec:results} evaluates accuracy, design choices, detector coverage, and computational performance. Section~\ref{sec:conclusion} concludes the paper and discusses the remaining deployment limitations.

\section{Related Work}
\label{sec:related_work}

Head pose estimation has been studied with a range of sensing modalities, including single camera RGB images, depth sensors, multiple camera systems, and auxiliary sensing setups such as methods assisted by IMUs or infrared sensors~\cite{Asp23,Aba22}. Among these, single camera RGB input remains the most practical setting for low cost and widely deployable systems, since additional sensors or camera setups increase hardware complexity and are less aligned with microcontroller class edge deployment. In this context, prior work is most relevant along two lines: RGB image based methods, which estimate pose directly from a single image and often achieve strong accuracy, and landmark based methods, which operate on sparse facial keypoints and are more suitable for efficient inference. For TinyML, the main challenge is not only predictive accuracy, but also low memory use, low computational cost, and compact input representations.

\subsection{RGB Image Based Methods}

Single camera RGB image based head pose estimation is dominated by CNN and transformer models. Early methods such as Biternion Nets~\cite{Bey15}, HopeNet~\cite{Fine-Grained}, and FSA-Net~\cite{fsanet} showed that pose can be predicted directly from face images without explicit geometric fitting. Later work improved robustness through stronger backbones and better output representations, including WHENet~\cite{whenet}, 6DRepNet~\cite{6drepent}, HeadPosr~\cite{headposr}, and TokenHPE~\cite{tokenhpe}. Most of these methods predict Euler angles directly, while some recent studies use quaternions, vector based rotation cues, or continuous 6D rotation representations to improve robustness for large or unconstrained poses~\cite{quatnet,vectorhpe,6drotation,fullrangehpe}. Although these methods benefit from rich visual information and often achieve strong accuracy, they still rely on dense image processing and are generally less suitable for strict TinyML constraints~\cite{tflitemicro,mcunet}.

\begin{figure*}[!t]
\centering
\includegraphics[width=\linewidth]{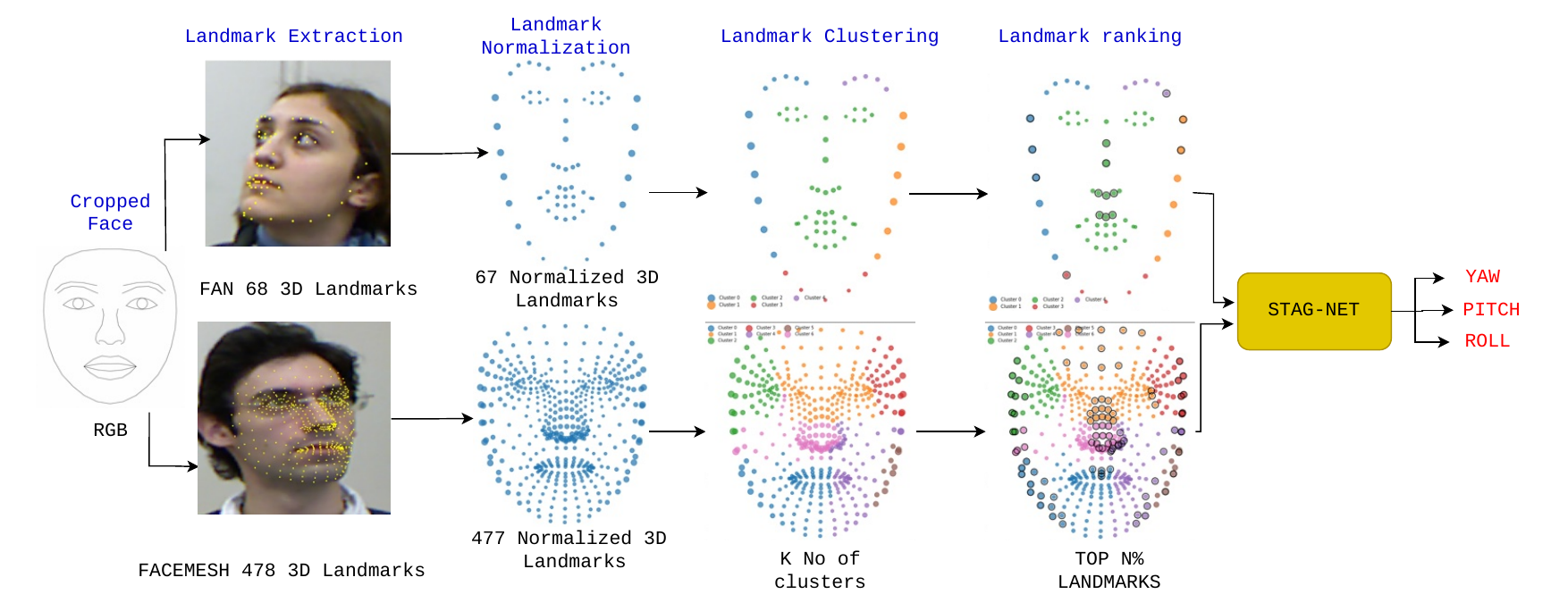}
\caption{Overview of the proposed head pose estimation framework. Facial landmarks are first extracted from a cropped facial image and subsequently normalized with respect to the nose point by removing the absolute nose coordinates. The remaining landmarks are then grouped into spatially coherent clusters, ranked according to their motion characteristics, and filtered using a top $N\%$ landmark selection strategy. The selected subset of landmarks is finally processed by STAG-Net to estimate the yaw, pitch, and roll angles of the head.}
\label{fig:overview_diagram}
\end{figure*}

\subsection{Landmark Based Methods}

Landmark based methods estimate pose from sparse 2D or 3D facial keypoints rather than raw pixels. Classical methods used geometric formulations such as point correspondence and perspective n point style fitting~\cite{3dposes}, while later methods used learned regressors on top of landmark coordinates~\cite{Wer17,Der17}. Geometry based approaches such as~\cite{Yuan} further showed that reasonable pose estimates can be obtained from only a few facial points at very low cost. This makes landmark based estimation especially attractive for embedded deployment.

However, many landmark methods represent facial points as fixed order vectors and use simple regression models, which limits their ability to capture structural relations among landmarks~\cite{Wer17,Der17}. More recent graph based methods, including EVA-GCN~\cite{eva}, OsGG-Net~\cite{osgnet}, and~\cite{singlegnn}, address this by explicitly modeling interactions among facial points. Set based approaches are still rarer. Deep Sets~\cite{deepsets}, PointNet~\cite{pointnet}, PointNet++~\cite{pointnet++}, and Set Transformer~\cite{settransformer} provide the main permutation invariant learning tools, but direct application to head pose remains limited. DS-HPE~\cite{ds_hpe} and GADS~\cite{gads} are notable examples. Even so, lightweight landmark native models that jointly capture local facial structure and cross region dependencies remain scarce.

\subsection{Efficient Models and TinyML Approaches}

A parallel line of work has focused on compact and deployable models through lightweight architectures, pruning, and quantization. MobileNet~\cite{mobilenet} and EfficientNet~\cite{efficient} are standard compact CNN families, and head pose models such as WHENet~\cite{whenet}, LwPosr~\cite{lwposr}, and~\cite{accuratehpe} adopt similar efficiency goals. At the deployment level, TensorFlow Lite Micro~\cite{tflitemicro}, CMSIS-NN~\cite{cmsis}, MCUNet~\cite{mcunet}, and MCUNetV2~\cite{mcunetv2} show that TinyML feasibility depends not only on parameter count, but also on SRAM, flash, and operator support. Quantization and pruning methods further improve deployability~\cite{memorydriven,pruning_hpe,diff_pruning}. However, most efficient head pose studies still compress RGB pipelines rather than starting from compact landmark inputs, leaving room for structure aware landmark models that preserve the efficiency of landmark representations while improving relational modeling for real time TinyML deployment. These limitations motivate the grouped landmark representation introduced in the following section.

\section{Our Method}
\label{sec:method}
\begin{figure*}[!t]
    \centering
    \begin{subfigure}{0.155\textwidth}
        \includegraphics[width=\linewidth]{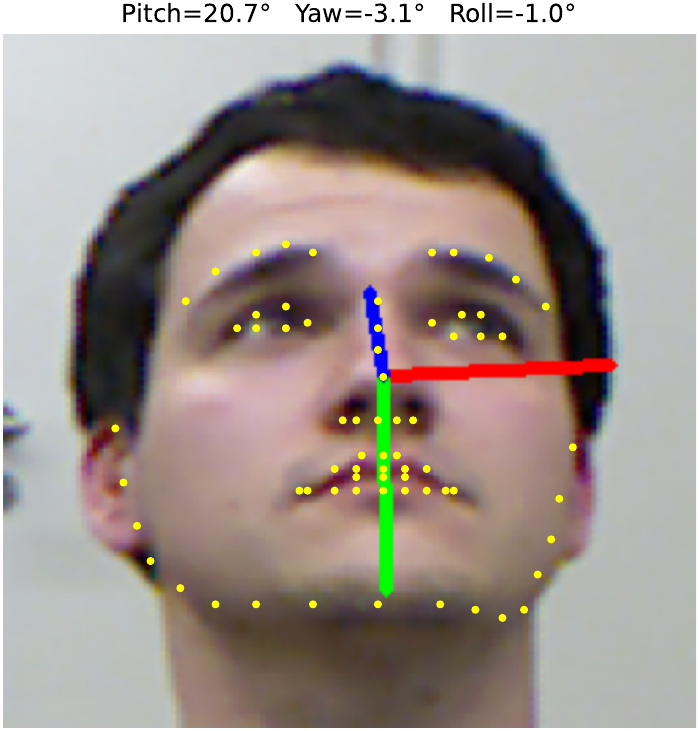}
    \end{subfigure}\hfill
    \begin{subfigure}{0.155\textwidth}
        \includegraphics[width=\linewidth]{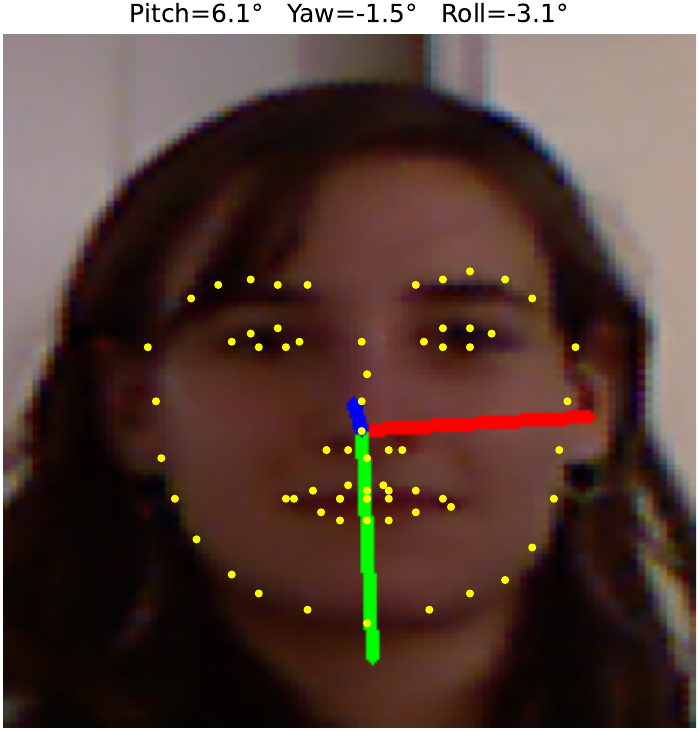}
    \end{subfigure}\hfill
    \begin{subfigure}{0.155\textwidth}
        \includegraphics[width=\linewidth]{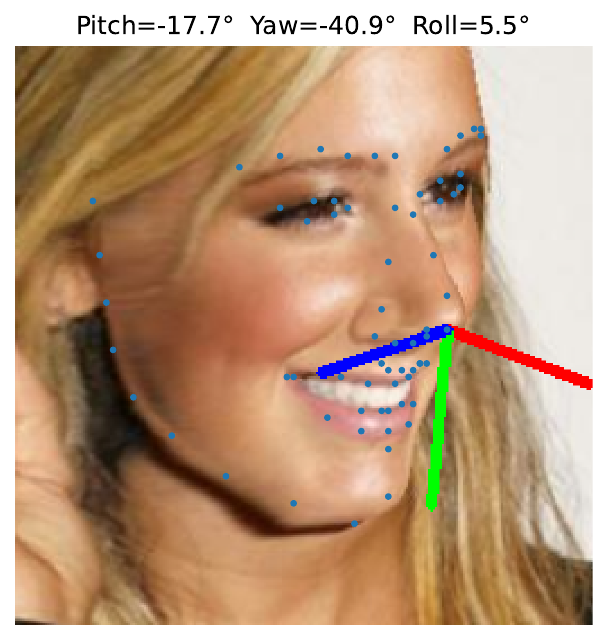}
    \end{subfigure}\hfill
    \begin{subfigure}{0.155\textwidth}
        \includegraphics[width=\linewidth]{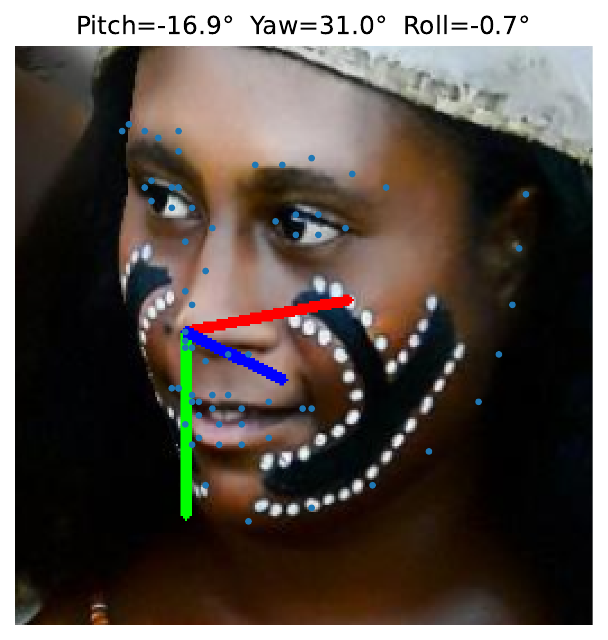}
    \end{subfigure}\hfill
    \begin{subfigure}{0.155\textwidth}
        \includegraphics[width=\linewidth]{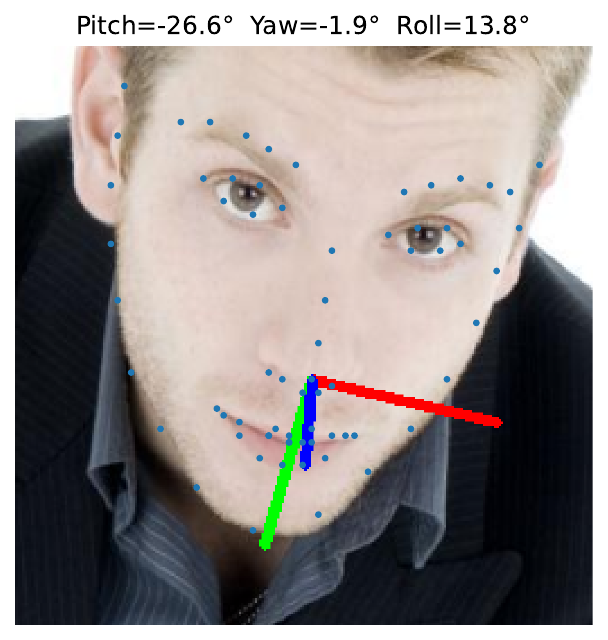}
    \end{subfigure}\hfill
    \begin{subfigure}{0.155\textwidth}
        \includegraphics[width=\linewidth]{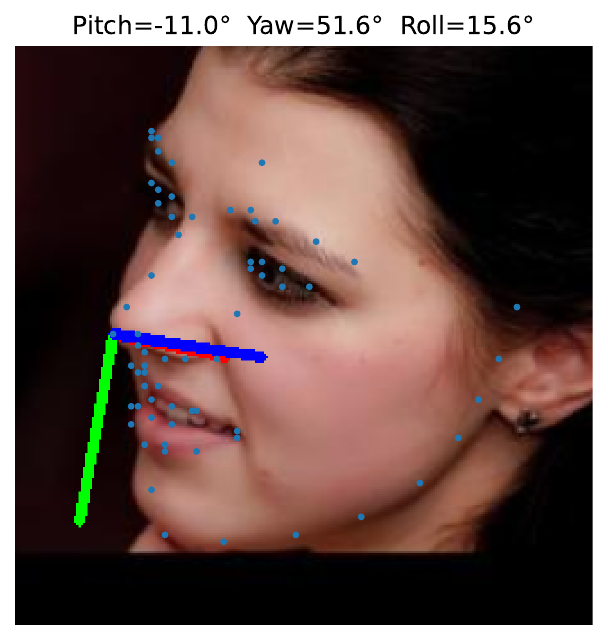}
    \end{subfigure}

    \caption{Visualization of 68 3D facial landmarks extracted using the Face Alignment Network (FAN), along with the corresponding Euler angles.}
    \label{fig:fan_landmarks}
\end{figure*}

\begin{figure*}[!t]
    \centering

    \begin{subfigure}{0.155\textwidth}
        \includegraphics[width=\linewidth]{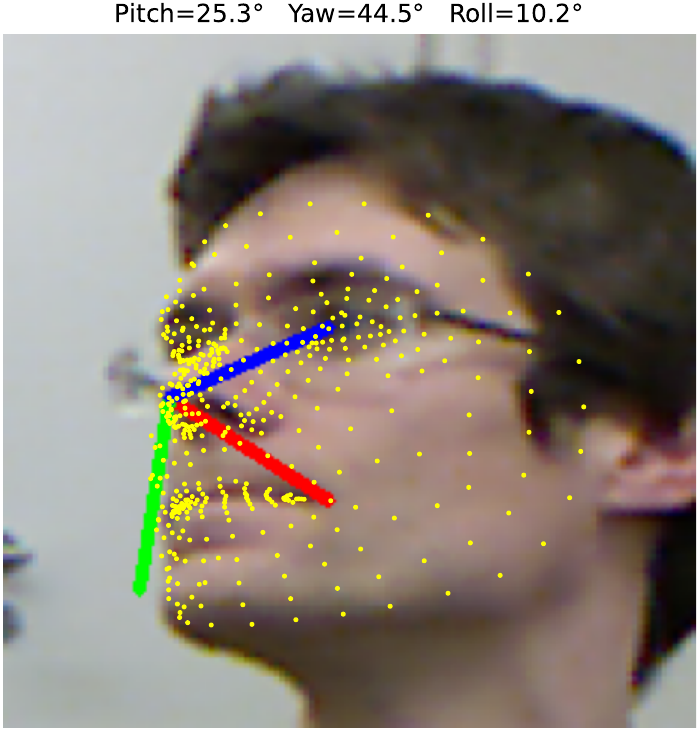}
    \end{subfigure}\hfill
    \begin{subfigure}{0.155\textwidth}
        \includegraphics[width=\linewidth]{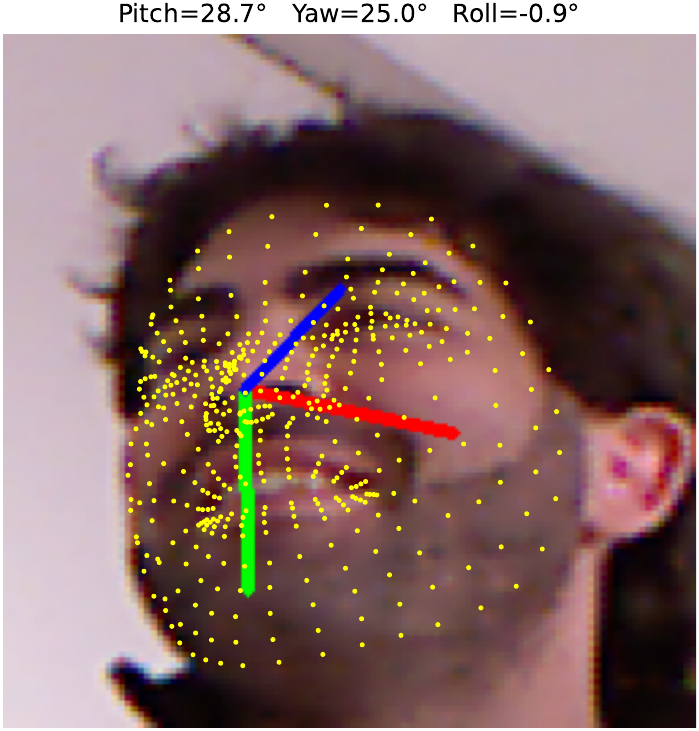}
    \end{subfigure}\hfill
    \begin{subfigure}{0.155\textwidth}
        \includegraphics[width=\linewidth]{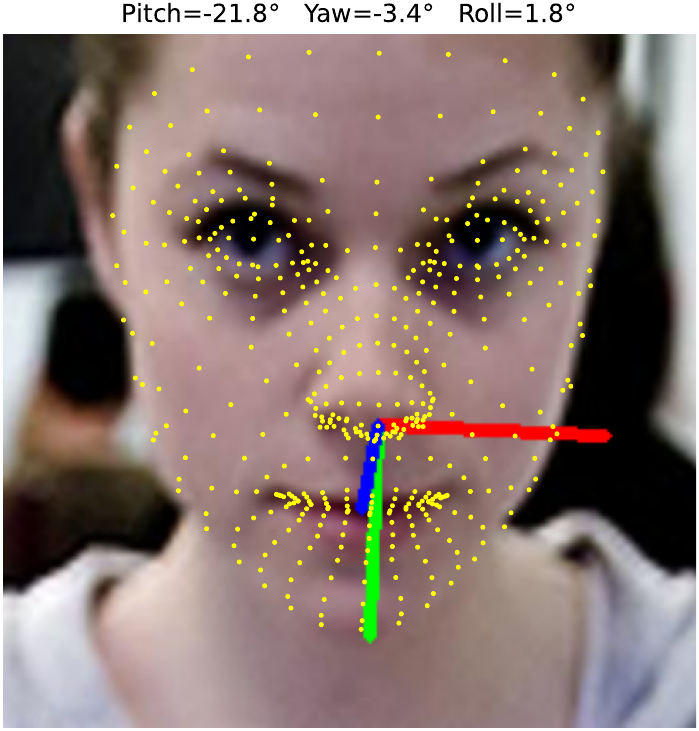}
    \end{subfigure}\hfill
    \begin{subfigure}{0.155\textwidth}
        \includegraphics[width=\linewidth]{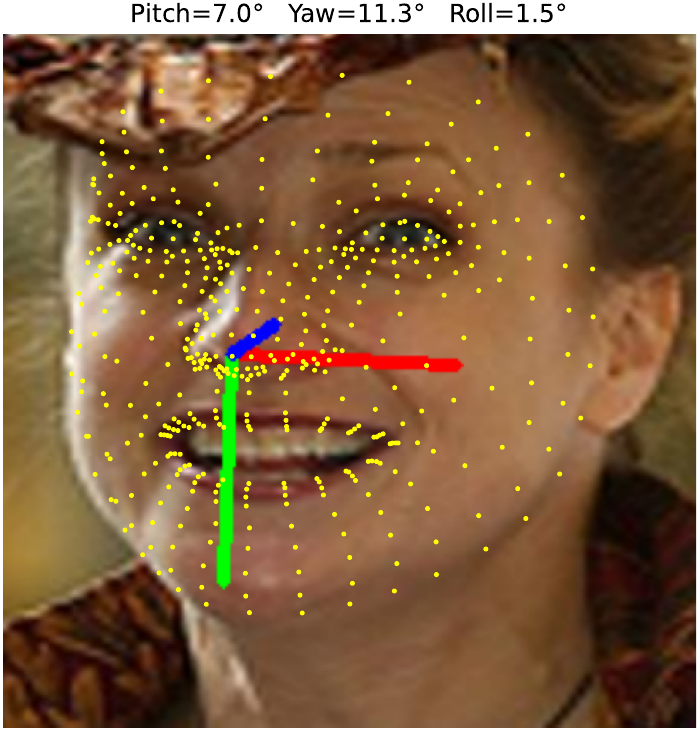}
    \end{subfigure}\hfill
    \begin{subfigure}{0.155\textwidth}
        \includegraphics[width=\linewidth]{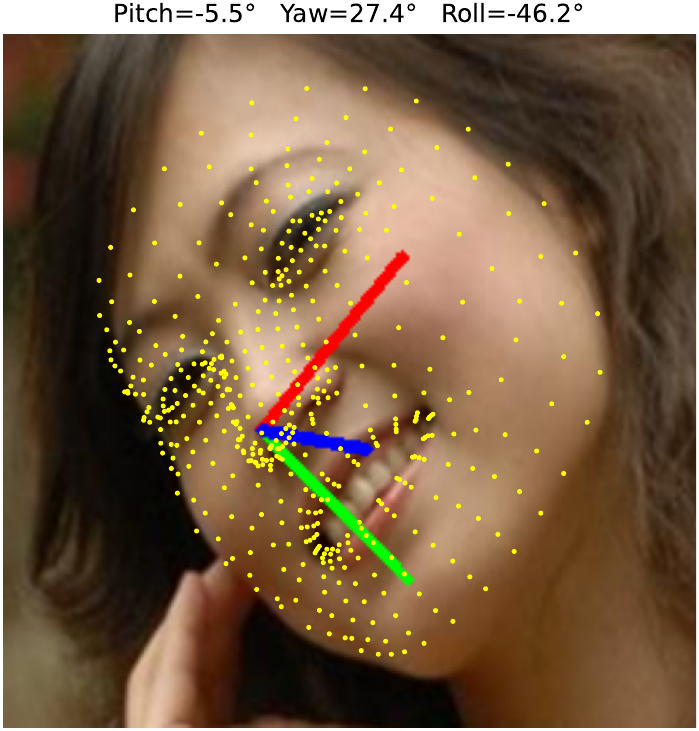}
    \end{subfigure}\hfill
    \begin{subfigure}{0.155\textwidth}
        \includegraphics[width=\linewidth]{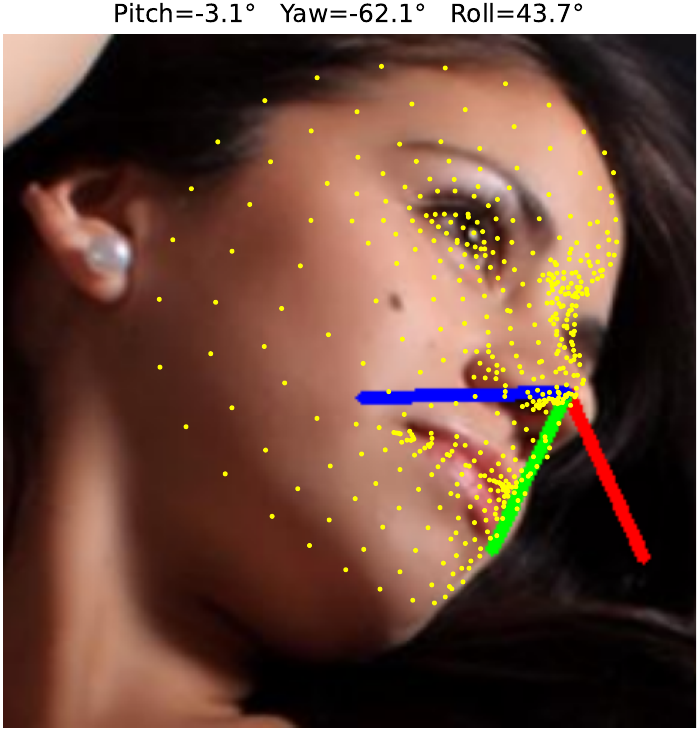}
    \end{subfigure}
    \caption{Visualization of 478 3D facial landmarks extracted using FaceMesh, along with the corresponding Euler angles.}
    \label{fig:mp_landmarks}
\end{figure*}
This section presents STAG-Net and follows the data flow shown in Fig.~\ref{fig:overview_diagram}. Facial landmarks are first extracted from a cropped face image and normalized to reduce translation and scale variation. The normalized landmarks are then grouped into spatially coherent regions, after which movement statistics are used to retain the most pose sensitive landmarks from each region. STAG-Net processes every group as an unordered set through an independent DeepSet encoder with a common architecture. The resulting group tokens interact through self attention and are combined with a compact geometry descriptor before the network predicts yaw, pitch, and roll. The method extends the DeepSet~\cite{deepsets} representation used in DS-HPE~\cite{ds_hpe} by introducing automatic grouping, movement aware selection, attention across groups, and explicit global geometry. This sequence connects the compact landmark input directly to the accuracy and efficiency objectives established in the introduction.

\subsection{Problem Formulation}
Head pose estimation aims to predict the 3D orientation of a human head from visual or geometric input. In this work, we focus on the estimation of head pose based on landmarks, where the input is a set of detected facial landmarks. Formally, let $X = \{x_1,\ldots,x_N\}$, where each landmark $x_i \in \mathbb{R}^3$ represents the 3D spatial coordinates of a facial keypoint. The goal is to learn a mapping
\begin{equation}
f: \mathbb{R}^{N \times 3} \rightarrow \mathbb{R}^3
\end{equation}
that predicts the head pose in terms of yaw, pitch, and roll angles.

A key property of landmark data is that it naturally forms an unordered set, meaning that the prediction should be invariant to permutations of the input ordering. However, many existing approaches treat landmarks as fixed order vectors, which limits their ability to generalize and capture structural relationships.

In this work, we adopt the Euler angle representation for head pose due to its simplicity and interpretability, as well as its widespread use in prior literature. While alternative representations such as quaternions and continuous rotation matrices can provide improved numerical stability, Euler angles remain suitable for lightweight regression tasks and enable direct comparison with existing methods.

Given these considerations, the objective is to design a function $f(\cdot)$ that is both permutation invariant and computationally efficient, while effectively capturing the structural relationships among facial landmarks for accurate head pose estimation.

In the following subsections, we describe the proposed STAG-Net pipeline in detail, including input representation, preprocessing, and the architectural components of the model.

\subsection{Input Representation and Preprocessing}
\label{sec:preprocessing}

\begin{figure}[!t]
    \centering
\includegraphics[width=\linewidth]{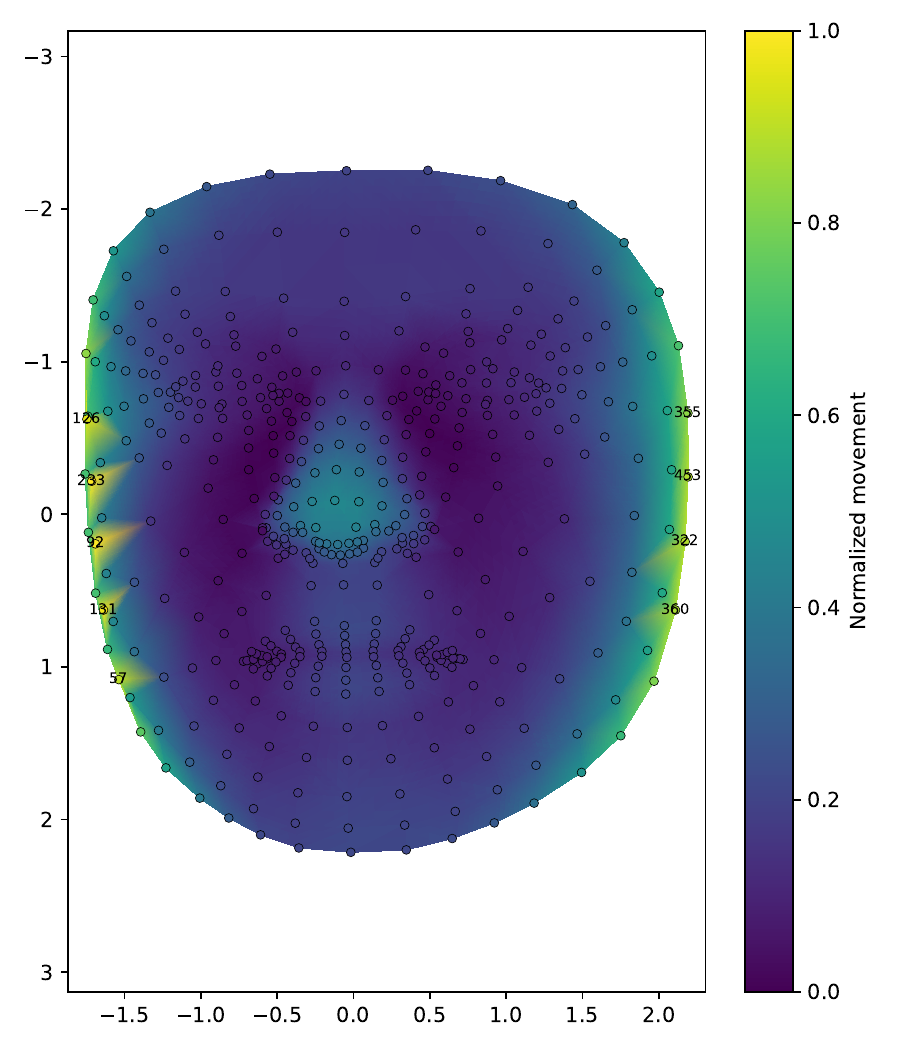}
\caption{Movement heatmap for landmarks detected by FaceMesh.}
    \label{fig:mp_heatmap}

\end{figure}

\begin{figure}[!t]
    \centering
\includegraphics[width=\linewidth]{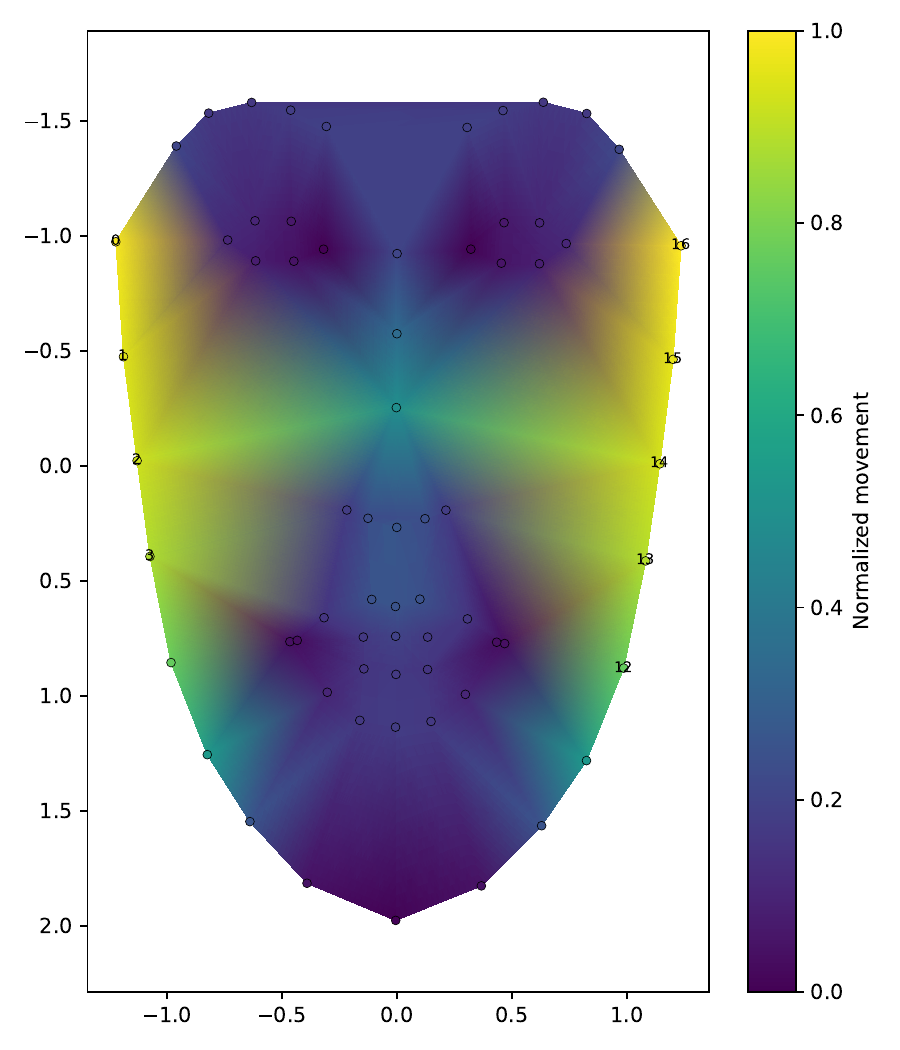}
\caption{Movement heatmap for landmarks detected by FAN.}
    \label{fig:fan_heatmap}
\end{figure}

Facial landmarks used in this work are obtained using two widely
adopted landmark detectors, namely the \textit{Face Alignment
Network} (FAN)~\cite{howfar} and \textit{FaceMesh}~\cite{mediapipe},
which is distributed as part of Google's MediaPipe framework. The
landmarks are illustrated in the first stage of
Fig.~\ref{fig:overview_diagram}. These detectors provide reliable 3D
landmark estimates under varying pose and lighting conditions, while
maintaining a balance between accuracy and computational efficiency.

Specifically, \textit{FAN} produces 68 3D facial landmarks, whereas \textit{FaceMesh} produces 478 3D landmarks. In this work, we utilize the 3D coordinates $(x, y, z)$ extracted from these detectors, enabling the model to capture richer geometric information than purely 2D representations. Figs.~\ref{fig:fan_landmarks} and~\ref{fig:mp_landmarks} present the landmark distributions generated by the two detectors on randomly selected face images from the three datasets, along with their corresponding ground truth yaw, pitch, and roll annotations.

The detected landmarks are represented as a set $X = \{x_1,\ldots,x_N\}$, where each $x_i \in \mathbb{R}^3$. Since the ordering of landmarks does not carry semantic meaning for the learning process, the input is treated as an unordered set, motivating the use of permutation invariant set modeling~\cite{deepsets}.
To enhance robustness and support effective learning, we apply a normalization step to the input landmarks. Specifically, the landmarks are centered with respect to a reference point, chosen as the nose landmark (i.e., the 30th landmark for FAN and the 0th landmark for FaceMesh). This procedure removes global translation effects and ensures that the model concentrates on the relative geometric configuration.

Following centering, each coordinate dimension is scaled by its range within the sample to reduce variation caused by face size and camera distance. For coordinate dimension $d\in\{x,y,z\}$, the normalized value is
\begin{equation}
\tilde{x}_{i,d} =
\frac{x_{i,d}-x_{\mathrm{ref},d}}
{\max_j x_{j,d}-\min_j x_{j,d}+\epsilon},
\end{equation}
where $x_{\mathrm{ref},d}$ is the reference coordinate and $\epsilon$ prevents division by zero.

Finally, the reference landmark is removed from the input representation, resulting in 67 processed landmarks for FAN and 477 processed landmarks for FaceMesh. Stage 2 of Fig.~\ref{fig:overview_diagram} illustrates the normalized landmark distributions for both landmark detectors over their corresponding mean face shapes. The nose tip landmark, which serves as the reference point for centering, has been removed from both landmark configurations and is therefore not shown in the visualizations.

\subsection{Movement Aware Landmark Grouping and Selection}

\begin{figure}[!t]
    \centering
    \includegraphics[width=\columnwidth]{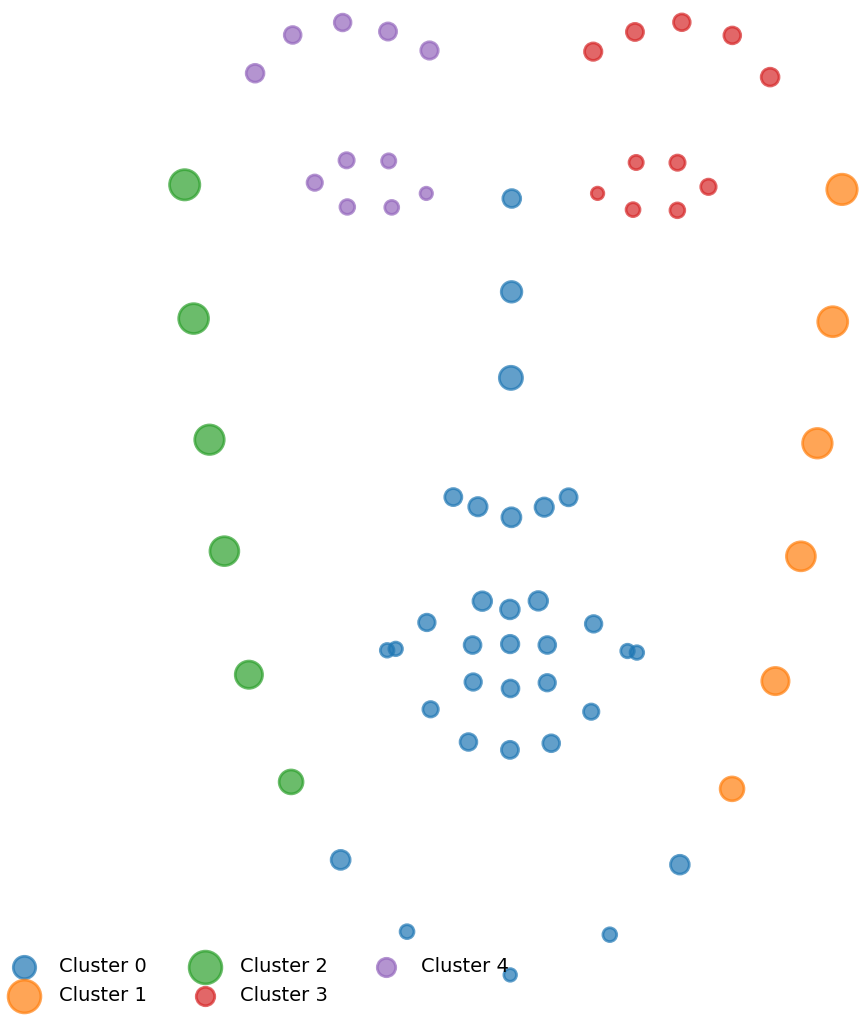}
   \caption{Spatial landmark clusters obtained from FAN landmarks.}
    \label{fig:fan_clusters}
\end{figure}

\begin{figure}[!t]
    \centering
\includegraphics[width=\linewidth]{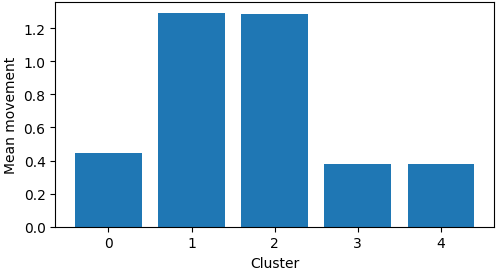}
\caption{Average movement magnitude within each landmark cluster.}
    \label{fig:fan_movements}
\end{figure}

\begin{figure}[!t]
    \centering
\includegraphics[width=\linewidth]{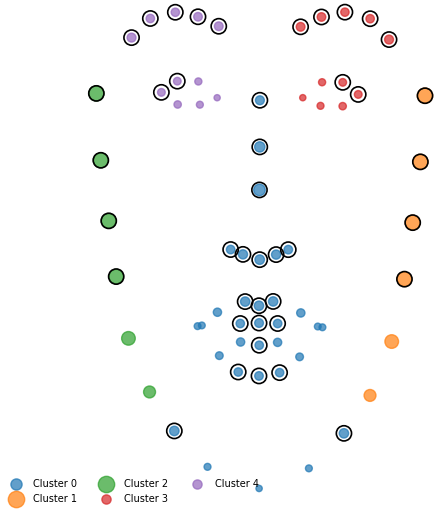}
\caption{Landmarks retained from each cluster using the top $N\%$ selection rule.}
    \label{fig:fan_clusters_selected}
\end{figure}

Beyond landmark normalization, the proposed framework performs a movement aware landmark grouping and selection procedure to identify compact yet informative landmark subsets. The procedure consists of three main stages: (i) movement analysis, (ii) landmark clustering, and (iii) cluster wise landmark selection.

In the first stage, landmark movement statistics are computed from the normalized landmarks to quantify the relative sensitivity of each landmark to head pose variations. Let the normalized coordinates of landmark \(i\) under pose sample \(j\) be denoted by \(\mathbf{p}_{ij}\). The average movement magnitude of landmark \(i\) is computed as

\begin{equation}
m_i=\frac{1}{M}\sum_{j=1}^{M}\left\|\mathbf{p}_{ij}-\bar{\mathbf{p}}_i\right\|_2,
\end{equation}

where \(M\) denotes the number of pose samples and \(\bar{\mathbf{p}}_i\) is the mean location of landmark \(i\) across all samples. Figs.~\ref{fig:fan_heatmap} and~\ref{fig:mp_heatmap} visualize the resulting landmark movement heatmaps. It can be observed that landmark motion is not uniformly distributed across the face, with facial contour and boundary landmarks generally exhibiting larger pose related displacements than central facial landmarks. These heatmaps provide a qualitative indication of which landmarks are likely to contain richer pose related information, motivating the subsequent grouping and movement based landmark selection procedure. These movement statistics are computed once on the training split only; the resulting cluster assignments and selected landmark indices are then fixed and reused unchanged at test time, avoiding any test set leakage.

In the second stage, the training set mean positions of the normalized landmarks are partitioned into spatially coherent facial regions using connectivity constrained agglomerative hierarchical clustering. The spatial representation used for clustering is
\begin{equation}
\mathbf{X}_{xy}=\{\bar{\mathbf{p}}_1^{xy},\bar{\mathbf{p}}_2^{xy},\ldots,\bar{\mathbf{p}}_N^{xy}\},
\end{equation}
where \(N\) denotes the number of facial landmarks and \(\bar{\mathbf{p}}_i^{xy} \in \mathbb{R}^2\) is the frontal \((x,y)\) projection of the mean normalized position of landmark \(i\) over the training split. The clustering algorithm groups neighboring landmarks while preserving the underlying facial topology. The optimal number of clusters is determined using silhouette analysis, enabling the grouping strategy to adapt automatically to different facial landmark detectors without requiring manually defined facial regions. Fig.~\ref{fig:fan_clusters} illustrates the resulting landmark clusters obtained using the FAN landmark detector. The effect of varying the number of neighbors in the connectivity graph is investigated in the design analysis.

Following clustering, the movement characteristics of each landmark group are analyzed by computing the average movement magnitude within every cluster,
\begin{equation}
\mu_k=\frac{1}{|C_k|}\sum_{i\in C_k}m_i,
\end{equation}
where \(C_k\) denotes the set of landmarks belonging to cluster \(k\). The corresponding cluster wise movement statistics are presented in Fig.~\ref{fig:fan_movements}. The results show that different facial regions contribute unequally to head pose estimation, with certain clusters exhibiting substantially larger movement magnitudes than others. This observation validates the use of cluster wise analysis for identifying informative facial regions while preserving structural locality.

Finally, landmarks within each cluster are ranked according to their movement magnitude. For a cluster containing \(|C_k|\) landmarks, the number of retained landmarks is determined as

\begin{equation}
n_k=\left\lceil\frac{N_s}{100}|C_k|\right\rceil,
\end{equation}

where \(N_s\) denotes the selected percentage of landmarks. The indices of the \(n_k\) landmarks with the highest movement magnitudes are retained,

\begin{equation}
S_k=\operatorname{TopIndices}_{n_k}\left(\{m_i\}_{i\in C_k}\right),
\end{equation}

and the selected landmark group is \(\widetilde{C}_k=\{x_i\mid i\in S_k\}\). This preserves informative landmarks from every facial region while reducing the input dimensionality.

Fig.~\ref{fig:fan_clusters_selected} highlights the landmarks retained within each discovered cluster after applying the proposed movement based selection strategy. It can be observed that informative landmarks are distributed across all facial regions rather than being concentrated within a single area, demonstrating that the proposed selection method preserves the structural diversity of the face while eliminating less informative landmarks.

The complete landmark grouping and selection procedure is illustrated in Fig.~\ref{fig:overview_diagram}. For both FAN and FaceMesh, the figure presents the normalized landmark configuration, the automatically discovered landmark clusters, and the final subset of selected landmarks obtained through movement based ranking. The resulting landmark groups form the input to the proposed STAG-Net architecture.

\subsection{STAG-Net Architecture}

\begin{figure*}[!t]   
\centering
\begin{subfigure}[t]{0.4\linewidth}
    \centering
    \includegraphics[width=\linewidth]{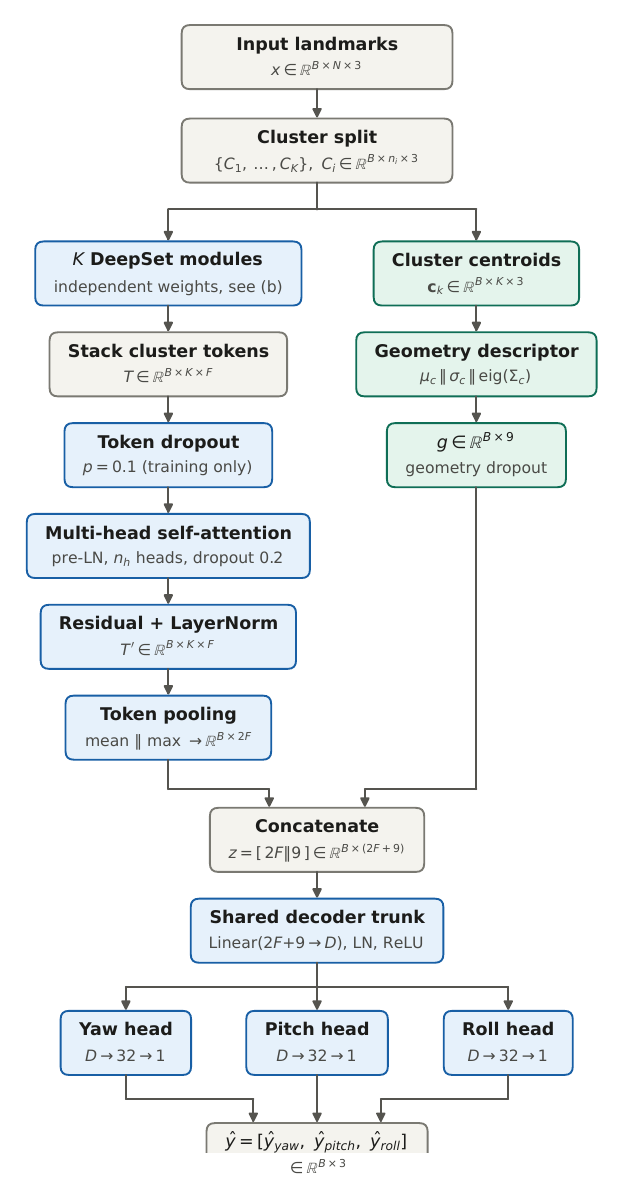}
    \caption{STAG-Net system architecture.}
\end{subfigure}
\hfill
\begin{subfigure}[t]{0.4\linewidth}
    \centering
    \includegraphics[width=\linewidth]{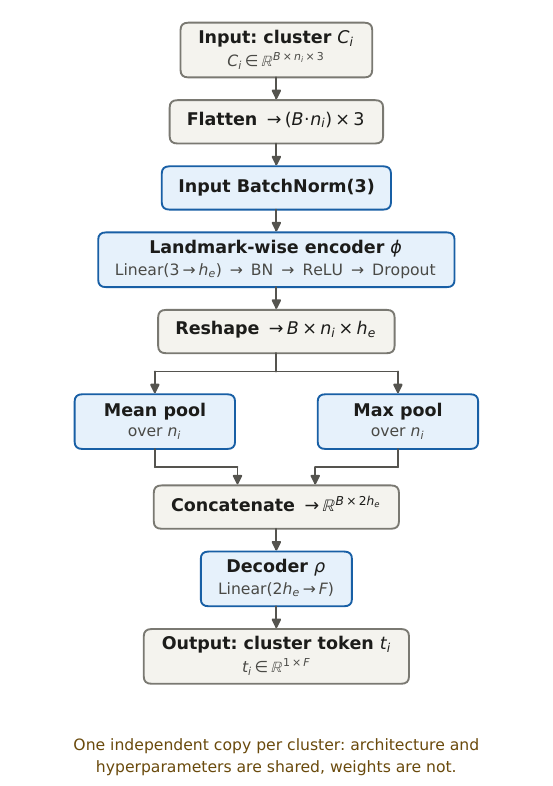}
    \caption{Per cluster DeepSet module.}
\end{subfigure}
\caption{Overview of the proposed STAG-Net architecture (a) and the per cluster DeepSet module used for cluster tokenization (b). All DeepSet modules share the same architecture and hyperparameters but maintain independent weights.}
\label{fig:stagnet}
\end{figure*}
\subsubsection{STAG-Net Overview}

The overall architecture of the proposed Structure Aware TinyML Attention over Groups Network (STAG-Net) is illustrated in Fig.~\ref{fig:stagnet}. Following the movement aware landmark grouping and selection procedure described in the previous subsection, the selected landmark groups are provided as the input to STAG-Net. Unlike conventional landmark based methods that process all facial landmarks jointly, the proposed framework models each landmark group independently while explicitly learning interactions between different facial regions.

Each landmark group is first encoded by a dedicated DeepSet module into a compact token representation. All modules share the same architecture and hyperparameters but maintain independent weights, keeping the design lightweight while allowing each module to specialize to the geometry of its facial region. The resulting group tokens are subsequently processed using an inter group multihead self attention module, enabling the network to capture dependencies between different facial regions that cannot be modeled by individual landmark groups alone.

In parallel, a compact geometry branch extracts lightweight statistical descriptors from the spatial arrangement of the landmark groups, providing complementary global geometric information. The attention features and geometry descriptors are fused and passed through a shared decoder trunk followed by lightweight per angle regression heads to predict the yaw, pitch, and roll angles.

By combining permutation invariant set encoding, inter group attention, and compact geometric descriptors, STAG-Net effectively captures both local and global facial structure while maintaining the low computational complexity required for TinyML deployment.

The following subsections describe each component of STAG-Net in detail.

\subsubsection{Per Group DeepSet Token Encoders}

Following the movement aware grouping and selection procedure, the input to STAG-Net consists of a set of $K$ landmark groups,

\begin{equation}
\mathcal{G}=\{C_1,C_2,\ldots,C_K\},
\end{equation}

where each group $C_k$ contains a subset of normalized facial landmarks. Since the ordering of landmarks within a group carries no semantic meaning, each landmark group is modeled as an unordered set. To obtain a compact representation for every group while preserving permutation invariance, STAG-Net employs per group DeepSet encoders with a shared architecture, illustrated in Fig.~\ref{fig:stagnet}(b).

DeepSets~\cite{deepsets} provide a principled framework for learning from unordered sets by decomposing the mapping into an element wise transformation followed by a permutation invariant aggregation. A general DeepSet function can be expressed as
\begin{equation}
f(X)=\rho\left(\operatorname{Agg}\left(\{\phi(x_i)\mid x_i\in X\}\right)\right),
\end{equation}

where $\phi(\cdot)$ denotes a shared element wise encoder, $\operatorname{Agg}(\cdot)$ is a permutation invariant aggregation operator, and $\rho(\cdot)$ represents a decoder that maps the aggregated representation to the desired feature space.

In the proposed architecture, each landmark group is processed by its own DeepSet module. All $K$ modules share an identical architecture and identical hyperparameters, but maintain independent weights, allowing each module to specialize to the geometry of its facial region; consequently, the total parameter count scales linearly with the number of clusters $K$. For landmark group $k$, the corresponding token representation is obtained as

\begin{equation}
t_k=\rho\left(\operatorname{Agg}\left(\{\phi(x_i)\mid x_i\in C_k\}\right)\right),
\end{equation}

where $t_k\in\mathbb{R}^{F}$ denotes the feature token corresponding to landmark group $C_k$, and $F$ is the token embedding dimension.

Within each group, the encoder $\phi(\cdot)$ first standardizes the raw coordinates with an input batch normalization layer and then projects every landmark independently into a higher dimensional feature space through a linear layer followed by batch normalization, ReLU activation, and dropout. Because the encoder parameters are shared across all landmarks within a group, each DeepSet module remains extremely small, and the per group specialization is obtained without sacrificing the compactness required for TinyML deployment.

To aggregate landmark features within each group, a permutation invariant mean and max pooling strategy is adopted. Let

\begin{equation}
E_k=\{\phi(x_i)\mid x_i\in C_k\}
\end{equation}

denote the encoded landmark features of group $k$. The aggregated representation is computed as

\begin{equation}
\operatorname{Agg}(E_k)=\operatorname{Concat}\left(\operatorname{Mean}(E_k),\operatorname{Max}(E_k)\right).
\end{equation}

The mean pooling operation summarizes the overall geometric structure of the landmark group, whereas max pooling emphasizes the most discriminative local features. Their combination provides a richer representation than either pooling strategy alone while remaining computationally efficient.

Finally, the aggregated feature vector is processed by the decoder $\rho(\cdot)$, implemented as a single linear layer, to generate the final group token. Repeating this procedure for every landmark group produces the token set

\begin{equation}
T=
[t_1,t_2,\ldots,t_K],
\end{equation}

which forms the input to the subsequent inter group self attention module.

Unlike conventional landmark based approaches that directly process all facial landmarks jointly, the proposed tokenization strategy first summarizes each facial region independently before modeling interactions between regions. This hierarchical representation improves computational efficiency while preserving the structural information required for accurate head pose estimation.

\subsubsection{Inter Group Self Attention}
\label{sec:attention}

Although the per group DeepSet encoders effectively capture local geometric relationships within each landmark group, head pose estimation depends on the coordinated movement of multiple facial regions. For example, pose related variations observed around the jawline, nose, and eyes are often correlated. To model these global structural dependencies, STAG-Net employs an inter group multihead self attention module that enables information exchange between the group level tokens.

Let

\begin{equation}
T=[t_1,t_2,\ldots,t_K]\in\mathbb{R}^{K\times F},
\end{equation}

denote the token matrix produced by the per group DeepSet encoders, where each token corresponds to one landmark group and $F$ is the token embedding dimension.

Before computing attention, layer normalization is applied to improve optimization stability,

\begin{equation}
\hat{T}=\operatorname{LayerNorm}(T).
\end{equation}

The normalized tokens are projected into query, key, and value representations,

\begin{equation}
\mathbf{Q}=\hat{T}W_Q,\qquad
\mathbf{K}=\hat{T}W_K,\qquad
\mathbf{V}=\hat{T}W_V,
\end{equation}

where $W_Q$, $W_K$, and $W_V$ are learnable projection matrices.

The attention weights are computed using scaled dot product attention,

\begin{equation}
A=\operatorname{Softmax}\left(\frac{\mathbf{Q}\mathbf{K}^{\top}}{\sqrt{d}}\right),
\end{equation}

where $d$ denotes the dimensionality of each attention head. The refined token representations are then obtained as

\begin{equation}
T_{\mathrm{att}}
=
A\mathbf{V}.
\end{equation}

To facilitate optimization and preserve the original group representations, a residual connection followed by layer normalization is employed,

\begin{equation}
T'=\operatorname{LayerNorm}\left(T+T_{\mathrm{att}}\right).
\end{equation}

Multihead attention is adopted to enable the model to jointly capture multiple types of relationships between different facial regions. Each attention head learns complementary interactions among landmark groups, allowing STAG-Net to exploit both local and global structural dependencies that cannot be captured by independently processing each landmark group. Since each head operates on a partition of the token embedding with dimension $d=F/n_h$, the number of heads $n_h$ controls the granularity of these interactions without affecting the parameter count of the attention module.

During training, token dropout is applied before the attention operation to improve robustness and reduce overfitting. Rather than removing individual landmark features, token dropout randomly suppresses a subset of landmark group tokens while ensuring that at least one token remains available for attention. This encourages the model to learn robust representations that do not rely excessively on any single facial region.

To obtain a fixed length representation that is invariant to group ordering, the refined tokens are aggregated using permutation invariant mean and max pooling,

\begin{equation}
f_{\mathrm{att}}=\operatorname{Concat}\left(\operatorname{Mean}(T'),\operatorname{Max}(T')\right)\in\mathbb{R}^{2F},
\end{equation}

which serves as the learned structural representation of the face and is subsequently fused with the compact geometry descriptor for the final pose regression stage.

\subsubsection{Compact Geometry Branch}

While the inter group self attention module learns feature interactions between landmark groups, it does not explicitly preserve the global spatial arrangement of the facial geometry. To complement the learned token representations, STAG-Net incorporates a lightweight geometry branch that summarizes the spatial distribution of the landmark groups using compact statistical descriptors. Unlike the attention branch, which learns task specific feature representations, the geometry branch directly encodes the geometric configuration of the face, providing complementary structural information with negligible computational overhead.

For each landmark group \(C_k\), the centroid is computed as

\begin{equation}
c_k=\frac{1}{|C_k|}\sum_{x_i\in C_k}x_i,
\end{equation}

where \(|C_k|\) denotes the number of landmarks belonging to group \(k\). The resulting centroid set

\begin{equation}
\mathcal{C}=\{c_1,c_2,\ldots,c_K\}
\end{equation}

provides a compact representation of the spatial layout of the grouped facial landmarks.

Three statistical descriptors are subsequently extracted from the centroid set. The centroid mean,

\begin{equation}
\mu_c=\frac{1}{K}\sum_{k=1}^{K}c_k,
\end{equation}

captures the overall spatial location of the landmark groups. The centroid standard deviation,

\begin{equation}
\sigma_c=\sqrt{\frac{1}{K}\sum_{k=1}^{K}(c_k-\mu_c)^2},
\end{equation}

quantifies the spatial dispersion of the groups, while the covariance matrix

\begin{equation}
\Sigma_c=
\frac{1}{K-1}
\sum_{k=1}^{K}
(c_k-\mu_c)(c_k-\mu_c)^{\top}
\end{equation}

captures the correlation between the centroid coordinates. For numerical and scale stability, the covariance matrix is normalized by its trace prior to the eigendecomposition, $\tilde{\Sigma}_c=\Sigma_c/(\operatorname{tr}(\Sigma_c)+\epsilon)$, where $\epsilon$ prevents division by zero. The eigenvalues of the normalized covariance matrix,

\begin{equation}
\lambda=
\operatorname{eig}(\tilde{\Sigma}_c),
\end{equation}

provide a compact description of the principal geometric variation of the grouped facial structure.

The three descriptors are concatenated to form the geometry feature vector

\begin{equation}
g=
\operatorname{Concat}
(\mu_c,\sigma_c,\lambda),
\end{equation}

which summarizes the global spatial configuration of the grouped facial landmarks using only a small number of features.

Compared with directly processing all landmark coordinates, the proposed geometry branch provides a compact structural representation while introducing only minimal computational cost. Consequently, it complements the learned token representations obtained from the attention branch and enhances the ability of STAG-Net to model global facial geometry without compromising its suitability for TinyML deployment.

\subsubsection{Feature Fusion and Multioutput Pose Regression}

The refined token representations obtained from the inter group self attention module capture complementary relationships between different facial regions, while the compact geometry branch explicitly encodes the global spatial arrangement of the landmark groups. To exploit the strengths of both representations, STAG-Net combines the learned attention features with the geometry descriptor through feature fusion.

Let $f_{\mathrm{att}} \in \mathbb{R}^{2F}$ denote the pooled attention representation and let $g \in \mathbb{R}^{9}$ denote the geometry descriptor. The fused feature representation is obtained by concatenating the two,

\begin{equation}
z=\operatorname{Concat}(f_{\mathrm{att}},g)\in\mathbb{R}^{2F+9},
\end{equation}

where $z$ represents the final feature vector used for pose estimation.

The fused representation is subsequently processed by a shared decoder consisting of fully connected layers with layer normalization, ReLU activation, and dropout. The shared decoder learns a compact latent representation that captures information common to all three pose components while maintaining a lightweight architecture suitable for TinyML deployment.

Finally, the shared latent representation is passed to three lightweight per angle regression heads, one each for yaw, pitch, and roll; every head consists of a single hidden layer with layer normalization, ReLU activation, and dropout, followed by a linear output unit. The final pose prediction is given by

\begin{equation}
\hat{\mathbf{y}}=f_{\mathrm{reg}}\left(f_{\mathrm{shared}}(z)\right),
\end{equation}

where $f_{\mathrm{shared}}(\cdot)$ denotes the shared decoder, $f_{\mathrm{reg}}(\cdot)$ denotes the three per angle regression heads, and

\begin{equation}
\hat{\mathbf{y}}
=
\left[
\hat{y}_{\mathrm{yaw}},
\hat{y}_{\mathrm{pitch}},
\hat{y}_{\mathrm{roll}}
\right]^{\top}
\end{equation}

contains the predicted yaw, pitch, and roll angles, respectively.

Compared with employing three fully independent prediction networks, the proposed design shares the majority of the learned representation across all pose components in the common trunk, while the lightweight per angle heads retain the flexibility to specialize to each rotation axis. This significantly reduces the number of trainable parameters and computational complexity without sacrificing the ability to jointly estimate the three head pose angles. Consequently, the proposed feature fusion and shared trunk regression strategy contributes to the overall efficiency of STAG-Net, making it well suited for deployment on resource constrained TinyML platforms.

\section{Experimental Setup}
\label{sec:experimental_setup}

This section presents the experimental setup adopted to evaluate the proposed STAG-Net. We first describe the benchmark datasets and their characteristics, followed by the evaluation metrics and implementation details used throughout this study. The experiments are designed to provide a consistent and reproducible assessment of prediction accuracy and computational efficiency under the two established evaluation protocols.
\subsection{Datasets and Evaluation Protocols}
\label{evaluation_protocols}
To comprehensively evaluate the proposed STAG-Net framework, experiments are conducted on three widely used head pose estimation datasets: 300W-LP, AFLW2000, and BIWI. These datasets contain substantial variations in pose, illumination, expression, and acquisition conditions, making them suitable for evaluating both robustness and generalization.

\textbf{300W-LP}~\cite{faceAlignment} is a large scale synthetic head pose estimation dataset derived from the original 300W face alignment dataset~\cite{300faces}. It employs 3D face profiling and image meshing techniques to synthetically generate large out of plane pose variations. The original dataset contains 61,225 samples, which further expands to approximately 122,450 images after horizontal flipping augmentation. The dataset has become the standard training dataset for many head pose estimation methods due to its large pose diversity and scale.

\textbf{AFLW2000}~\cite{aflw2000} contains the first 2,000 images from the AFLW dataset annotated with 3D facial landmarks and head pose labels. The dataset includes significant variations in illumination, facial expression, occlusion, and head orientation, making it a widely used benchmark for cross dataset evaluation.

\textbf{BIWI}~\cite{biwi} is a benchmark RGB and depth head pose estimation dataset collected in a controlled indoor environment. The dataset contains 15,678 frames captured from 20 subjects across 24 video sequences. In addition to RGB images, aligned depth frames and accurate pose annotations are also provided. The dataset includes substantial head pose variations and is widely used for evaluating head pose estimation methods under subject independent settings. In this work, only single RGB frames are utilized. 

\begin{figure}[!t]
    \centering
    \begin{subfigure}{0.8\linewidth}
        \centering
        \includegraphics[width=\linewidth]{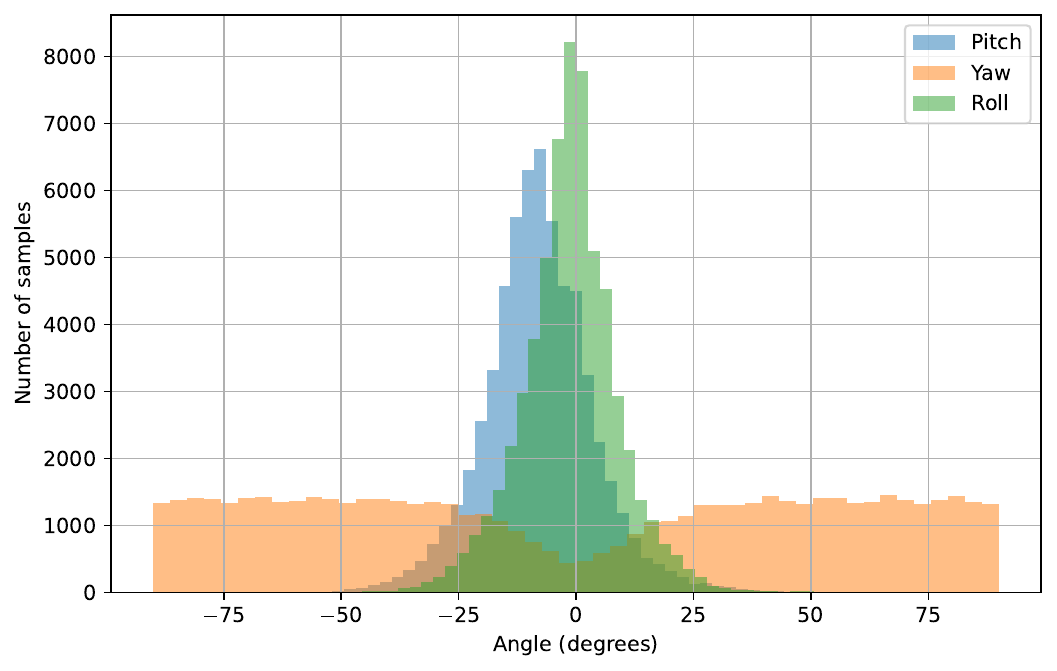}
        \caption{300W-LP}
    \end{subfigure}
    \begin{subfigure}{0.8\linewidth}
        \centering
        \includegraphics[width=\linewidth]{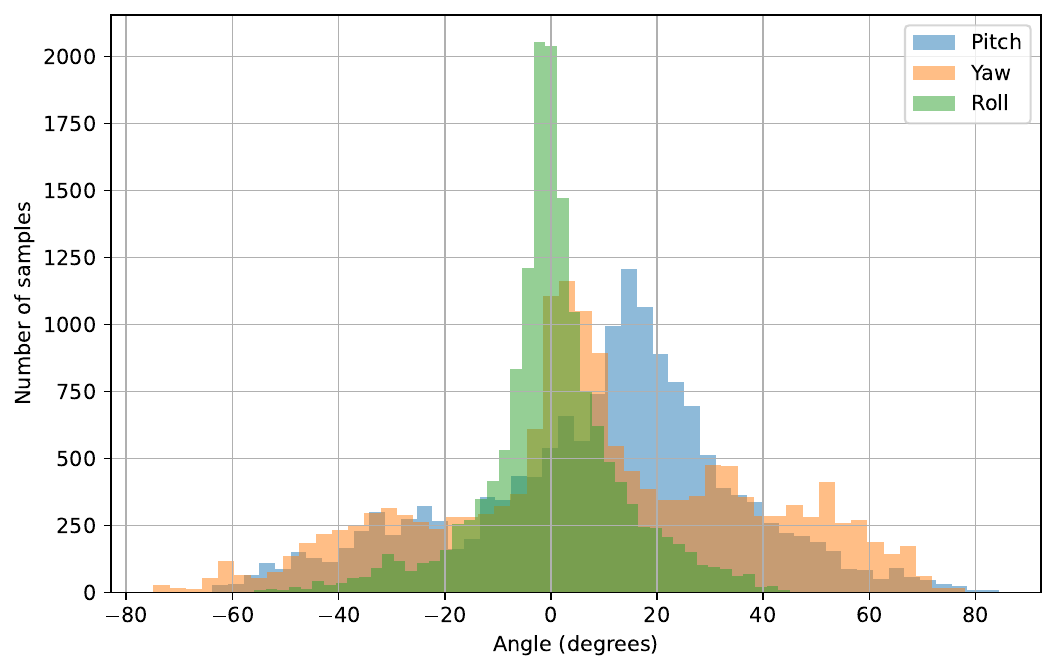}
        \caption{BIWI}
    \end{subfigure}
    \begin{subfigure}{0.8\linewidth}
        \centering
        \includegraphics[width=\linewidth]{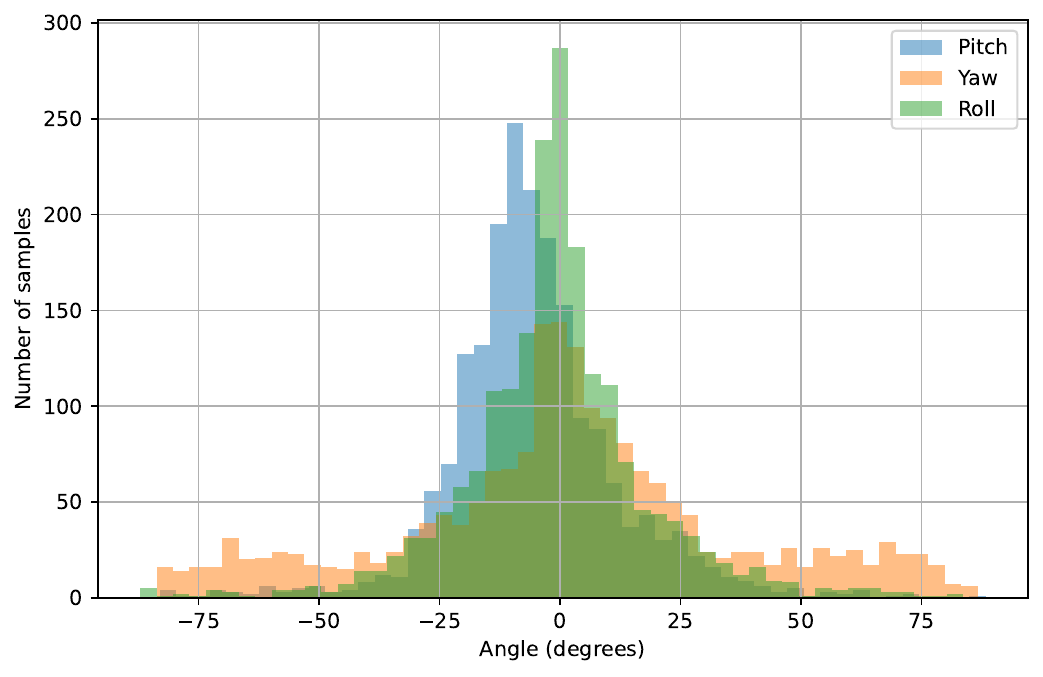}
        \caption{AFLW2000}
    \end{subfigure}
    \caption{Distribution of yaw, pitch, and roll angles for the three benchmark head pose estimation datasets.}
    \label{fig:angle_histograms}
\end{figure}

Since the distribution of head pose angles is rarely discussed in the
existing head pose estimation literature, we analyzed the Euler angle
distributions of the three benchmark datasets used in this study.
Fig.~\ref{fig:angle_histograms} presents the distributions of the
yaw, pitch, and roll angles for BIWI, 300W-LP, and AFLW2000, where
each histogram illustrates the number of samples corresponding to
different head pose angles. These distributions reveal the pose
diversity of each dataset and highlight the differences in angular
coverage among the three benchmarks. Table~\ref{tab:pose_ranges}
summarizes the exact minimum and maximum values of the three Euler
angles for each dataset. Notably, only 300W-LP spans the full
$\pm90^\circ$ yaw range, whereas BIWI covers a substantially narrower
angular envelope, particularly in yaw and roll; consequently, models
trained on 300W-LP under Protocol~I must generalize from a wider
synthetic pose distribution to the more constrained real world
distributions of the evaluation datasets.

During this analysis, we observed that the AFLW2000 dataset contains
36 samples with extremely large pose annotations that are
inconsistent with realistic head rotations. Specifically, these
samples exhibit pitch, yaw, and roll values ranging from
$-713.75^\circ$ to $893.75^\circ$, $-351.23^\circ$ to $187.79^\circ$,
and $-972.27^\circ$ to $334.00^\circ$, respectively. These values are
far beyond the physically plausible range of human head motion and
are likely the result of annotation errors or inconsistencies in the
dataset. Consequently, these anomalous samples were removed as a
data cleaning step prior to evaluation. After
removal, all AFLW2000 pose angles fall within
$[-83.67^\circ, 86.63^\circ]$ for yaw,
$[-82.96^\circ, 88.26^\circ]$ for pitch, and
$[-87.09^\circ, 83.66^\circ]$ for roll; these cleaned ranges are the
ones reported in Table~\ref{tab:pose_ranges}, and the histograms in
Fig.~\ref{fig:angle_histograms} likewise correspond to the cleaned
dataset. To the best of our knowledge, previous studies have not
explicitly reported the presence of these anomalous annotations or
documented their removal. We believe this observation provides useful
guidance for future researchers using the AFLW2000 dataset, as these
inconsistencies may influence both model training and evaluation if
left unaddressed.

\begin{table}[!t]
\centering
\caption{Head pose angle ranges of the datasets used in this study.
For AFLW2000, the reported ranges correspond to the cleaned dataset
after removal of the 36 anomalous samples described in the text.}
\label{tab:pose_ranges}
\small
\setlength{\tabcolsep}{3pt}
\begin{tabular}{lccc}
\toprule
\textbf{Dataset} & \textbf{Yaw ($^\circ$)} & \textbf{Pitch ($^\circ$)} & \textbf{Roll ($^\circ$)} \\
\midrule
BIWI     & $-74.89$ to $77.97$ & $-63.83$ to $84.37$ & $-64.70$ to $45.02$ \\
300W-LP  & $-90.00$ to $90.00$ & $-64.57$ to $61.90$ & $-73.43$ to $53.16$ \\
AFLW2000 & $-83.67$ to $86.63$ & $-82.96$ to $88.26$ & $-87.09$ to $83.66$ \\
\bottomrule
\end{tabular}
\end{table}

To make our experiments consistent, we reconstructed all three datasets from their original sources and used a common pose ordering of yaw, pitch, and roll. This step was necessary because public implementations differ in image resolution and label ordering. For example, the FSA-Net preprocessing pipeline uses $64\times64$ images, while landmark extraction is often performed at higher resolutions. Such differences can alter results when models are rerun. In our STAG-Net experiments, every image is resized to $224\times224$ before landmark extraction, and the same preprocessing and label conventions are used for both FAN and FaceMesh. Accuracy values quoted for prior methods in the comparison tables are taken from their cited publications under the corresponding protocols, while our runtime measurements use the official implementations on common hardware. For AFLW2000, our values are computed on the cleaned 1,964 sample subset, whereas values quoted from earlier publications may have been computed on the original 2,000 images. The AFLW2000 comparison should therefore be interpreted as contextual rather than as a strictly matched reevaluation of every baseline.

To follow established comparison settings, we adopt two commonly used evaluation protocols in the head pose estimation literature.

\paragraph{Protocol I: Cross Dataset Evaluation.}
Following FSA-Net~\cite{fsanet} and related work, the model is trained on the synthetic 300W-LP dataset and evaluated on the real world AFLW2000 and BIWI datasets. This protocol evaluates the generalizability of the model across different domains and acquisition conditions. 
For BIWI evaluation under this protocol, no tracking or temporal smoothing techniques are used. The proposed framework performs inference independently on each RGB frame.

\paragraph{Protocol II: Subject Independent BIWI Evaluation.}
Following TokenHPE~\cite{tokenhpe}, BIWI is divided at the sequence level into 16 training sequences and 8 test sequences, corresponding to two thirds of the sequences for training and one third for testing. This protocol evaluates the model's ability to generalize to the held out subjects and recording sessions defined by this split.

Although several methods evaluated under this protocol utilize additional modalities such as depth information or temporal sequences, the proposed STAG-Net framework operates solely on single RGB images and extracted 3D landmarks.

For both protocols, the STAG-Net regression error is computed on frames for which the selected landmark detector returns a valid landmark set. Frames without valid landmarks cannot be processed by a landmark based regressor. We therefore report the detector angle envelope in Table~\ref{tab:detector_coverage} and interpret MAE together with detector coverage. This distinction is particularly important at extreme poses, where the two detectors cover different subsets of the annotated range.

\subsection{Evaluation Metrics}

Performance is evaluated using the mean absolute error (MAE) of the
estimated Euler angles, which is the standard evaluation metric in the
head pose estimation literature. Let
$\Theta = \{\text{yaw}, \text{pitch}, \text{roll}\}$ denote the set of
pose angles, and let $y_i^{\theta}$ and $\hat{y}_i^{\theta}$ denote the
ground truth and predicted values of angle $\theta \in \Theta$ for test
sample $i$. The per angle MAE is computed as
\begin{equation}
\mathrm{MAE}_{\theta}
= \frac{1}{N}\sum_{i=1}^{N}
\bigl|\, y_i^{\theta} - \hat{y}_i^{\theta} \,\bigr|,
\label{eq:mae_angle}
\end{equation}
where $N$ is the number of test samples with valid landmark detections for the detector being evaluated. The overall MAE is then
obtained by averaging over the three pose angles,
\begin{equation}
\mathrm{MAE}
= \frac{1}{|\Theta|}\sum_{\theta \in \Theta}
\mathrm{MAE}_{\theta}.
\label{eq:mae_overall}
\end{equation}
Equation~\eqref{eq:mae_angle} produces the per angle columns reported
in Tables~\ref{tab:protocol1} and~\ref{tab:protocol2}, while
Equation~\eqref{eq:mae_overall} corresponds to their final MAE column.

\paragraph{Model Complexity Metrics}

In addition to prediction accuracy, the computational efficiency of the proposed framework is evaluated using the number of parameters and Floating Point Operations (FLOPs), which are commonly used metrics in TinyML and efficient deep learning research~\cite{mobilenet,efficient}.

Unlike conventional RGB based head pose estimation methods, the proposed STAG-Net operates on compact landmark representations and lightweight set based processing modules. Therefore, parameter count and FLOPs are particularly relevant for quantifying the efficiency gains achieved through the proposed architecture and assessing its suitability for deployment in resource constrained environments.

The number of trainable parameters provides a measure of the memory required to store the model. For a fully connected layer with $n_{in}$ input units and $n_{out}$ output units, the number of parameters is computed as:

\begin{equation}
P = n_{in} \times n_{out} + n_{out},
\end{equation}

where the first term corresponds to the weights and the second term corresponds to the bias parameters. The total number of parameters of a model is obtained by summing the parameters across all trainable layers:

\begin{equation}
P_{\text{total}} = \sum_{l=1}^{L} P_l,
\end{equation}

where $L$ denotes the total number of trainable layers.

The parameter count directly affects the flash memory requirements of embedded systems and is therefore an important consideration for TinyML deployment.

To quantify computational complexity, we report floating point operations (FLOPs), which estimate the number of arithmetic operations required during inference. For a fully connected layer, the FLOPs can be approximated as:

\begin{equation}
F = 2 \times n_{in} \times n_{out},
\end{equation}

where the factor of two accounts for multiplication and accumulation operations.

The total FLOPs of a model are computed as:

\begin{equation}
F_{\text{total}} = \sum_{l=1}^{L} F_l.
\end{equation}

FLOPs provide a hardware independent estimate of computational cost and are frequently used as a proxy for inference latency and energy consumption in resource constrained systems~\cite{benchmarkingTinyML}.

Together, MAE, parameter count, and FLOPs provide a comprehensive assessment of both predictive performance and deployment suitability. Since STAG-Net is primarily composed of fully connected layers and attention modules, the reported parameter counts and FLOPs closely reflect the actual memory and computational requirements of the deployed model.

Since TinyML deployment requires optimizing accuracy and resource cost
jointly rather than in isolation, we additionally assess the compared
methods through a Pareto dominance analysis over the joint
(MAE, parameter count) objective. A method is \emph{Pareto dominated}
if another method achieves both a lower or equal MAE and a lower or
equal parameter count, with strict improvement in at least one of the
two; the methods that are not dominated form the Pareto frontier of the
accuracy and efficiency tradeoff. This analysis, presented in
Section~\ref{performance_results}, identifies which methods represent
genuinely optimal operating points rather than merely favorable
positions along a single axis.

\subsection{Implementation Details}
\label{implementation_details}
Facial landmarks are extracted using the corresponding pretrained landmark detector, and the resulting coordinates are normalized as described in Section~\ref{sec:preprocessing} before they are provided to the network. The model is trained by minimizing the mean absolute error between the predicted and ground truth Euler angles using the settings summarized in Table~\ref{tab:implementation_details}. The architectural choices examined in the design analysis are reported in Section~\ref{sec:design}.

All experiments were conducted using the Google Colab environment. Performance comparisons were carried out on both the default Google Colab CPU runtime and the NVIDIA Tesla T4 GPU runtime. Model training and inference were evaluated on both platforms, and all reported execution times were measured under these hardware configurations. The proposed model was implemented using the PyTorch deep learning framework, and the primary training hyperparameters are summarized in Table~\ref{tab:implementation_details}.

\begin{table}[!t]
\centering
\caption{Implementation and training configuration.}
\label{tab:implementation_details}
\footnotesize
\begin{tabularx}{\columnwidth}{@{}>{\raggedright\arraybackslash}p{0.43\columnwidth}>{\raggedright\arraybackslash}X@{}}
\toprule
\textbf{Parameter} & \textbf{Value} \\
\midrule
Framework & PyTorch \\
Optimizer & Adam \\
Loss function & Mean absolute error (MAE) \\
Batch size & 128 \\
Training epochs & 150 \\
Initial learning rate & $5\times10^{-4}$ \\
Learning rate scheduler &
CosineAnnealingWarmRestarts \\

Initial restart period, \(T_0\) &
10 epochs \\

Restart period multiplier, \(T_{\mathrm{mult}}\) &
2 \\

Weight initialization & Xavier uniform \\
Dropout rate & 0.01 \\
Attention dropout & 0.2 \\
Token dropout & 0.1 \\

\bottomrule
\end{tabularx}
\end{table}

Prior to training, all face images were resized to a resolution of $224\times224$ pixels, after which facial landmarks were extracted using the corresponding landmark detector as part of the preprocessing pipeline. The extracted landmark coordinates were stored and subsequently used as the direct input to STAG-Net during training, ensuring a consistent preprocessing procedure across all experiments.

During training, no additional image enhancement techniques were used. Furthermore, early stopping was not utilized. Instead, the model achieving the lowest validation loss was checkpointed throughout training, and the best performing checkpoint was used for the final evaluation. This ensured a consistent model selection strategy while allowing all experiments to be trained for the predefined 150 epochs. The preprocessing, training, and evaluation procedures are kept identical across experiments so that differences can be attributed to the model configuration rather than to changes in the data pipeline. The following section reports the resulting accuracy and computational performance.

To support reproducibility, the implementation of STAG-Net is
publicly available at \url{https://github.com/asirigawesha/STAGNet}.

\section{Experimental Results}
\label{sec:results}

This section evaluates STAG-Net in the same order as the claims made in the introduction. We first compare prediction accuracy under the two protocols defined in Section~\ref{evaluation_protocols}. We then examine detector coverage, connectivity graph construction, landmark selection, architectural hyperparameters, and learning rate scheduling strategies. Finally, we analyze model size, regression FLOPs, landmark and regression pipeline throughput, training cost, and the joint relationship between accuracy and model size.

\subsection{Comparison with State of the Art Methods}

\begin{table*}[t]
\centering
\caption{Comparison with state of the art methods under Protocol~I,
grouped into RGB image based and landmark based methods, where \textbf{bold}
and \underline{underlined} values indicate the lowest and second lowest
reported numerical values per column across all methods. N/A indicates a value that could not be obtained.}
\label{tab:protocol1}
\resizebox{\textwidth}{!}{
\begin{tabular}{lccccccccc}
\toprule
\multirow{2}{*}{\textbf{Method}} &
\multirow{2}{*}{\textbf{Params (M)}} &
\multicolumn{4}{c}{\textbf{BIWI}} &
\multicolumn{4}{c}{\textbf{AFLW2000}} \\
\cmidrule(lr){3-6}\cmidrule(lr){7-10}
& & \textbf{Yaw} & \textbf{Pitch} & \textbf{Roll} & \textbf{MAE}
  & \textbf{Yaw} & \textbf{Pitch} & \textbf{Roll} & \textbf{MAE} \\
\midrule
\multicolumn{10}{l}{\textit{RGB image based methods}} \\
KEPLER (2017)\cite{kepler}              & N/R     & 8.80 & 17.30 & 16.20 & 13.90 & N/R   & N/R   & N/R   & N/R   \\
3DDFA (2019)\cite{3dffa}               & N/R     & 36.20 & 12.30 & 8.78 & 19.10 & 5.40 & 8.53 & 8.25 & 7.39 \\
TokenHPE (2023) \cite{tokenhpe}           & 86.42  & 4.51 & \textbf{3.95} & 2.71 & 3.72 & 5.54 & \textbf{3.36} & 4.08 & 4.66 \\
6DRepNet (2022) \cite{6drepent}           & 39.32  & \underline{3.23} & 5.32 & 2.78 & 3.78 & \textbf{3.27} & \underline{4.58} & \textbf{2.98} & \textbf{3.61} \\
TriNet (2021) \cite{triNet}             & 26.00  & 4.11 & 4.76 & 3.05 & 3.97 & 4.04 & 5.78 & 4.20 & 4.67 \\
Shao (2019) \cite{shao}             & 24.60  & 4.59 & 7.25 & 6.15 & 6.00 & 5.07 & 6.37 & 4.99 & 5.48 \\
HopeNet ($\alpha$=1) (2018) \cite{hopenet} & 23.90  & 4.81 & 6.61 & 3.27 & 4.90 & 6.92 & 6.64 & 5.67 & 6.41 \\
HopeNet ($\alpha$=2) (2018) \cite{hopenet} & 23.90  & 5.12 & 6.98 & 3.39 & 5.18 & 6.47 & 6.56 & 5.44 & 6.16 \\
WHENet (2020) \cite{whenet}       & 4.40   & 3.99 & \underline{4.39} & 3.06 & 3.81 & 5.11 & 6.24 & 4.92 & 5.42 \\
FSA-Caps-Fusion (2019)\cite{fsanet}     & 1.20   & 4.27 & 4.96 & 2.76 & 4.00 & 4.50 & 6.08 & 4.64 & 5.07 \\
SSR-Net-MD (2018) \cite{ssr}        & 0.20   & 4.49 & 6.31 & 3.61 & 4.65 & 5.14 & 7.09 & 5.89 & 6.01 \\
LwPosr (2022) \cite{lwposr}         & 0.15   & 4.11 & 4.87 & 3.19 & 4.05 & 4.80 & 6.38 & 4.88 & 5.35 \\
LwPosr-a (2022) \cite{lwposr} & 0.15   & 4.41 & 5.11 & 3.24 & 4.25 & 4.44 & 6.06 & 4.35 & 4.95 \\
\midrule
\multicolumn{10}{l}{\textit{Landmark based methods}} \\
FAN (12 points) (2017) \cite{howfar}  & 36.60  & 6.36 & 12.30 & 8.71 & 9.12 & 8.53 & 7.48 & 7.63 & 7.88 \\
Dlib (68 points) (2014) \cite{dlib}& 6.24   & 16.80 & 13.80 & 6.19 & 12.20 & 23.10 & 13.60 & 10.50 & 15.80 \\
EVA-GCN (2021) \cite{eva} & 3.30   & 4.46 & 5.34 & 4.11 & 4.64 & 4.01 & 4.78 & \textbf{2.98} & 3.92 \\
GADS (2025) \cite{gads}& 0.02   & 3.61 & 5.05 & 3.04 & 3.90 & 3.84 & 7.06 & 5.00 & 5.30 \\
Tourani et al.(2025) \cite{tourani2025unsupervised}& N/R  & 4.60 & 5.54 & 3.01 & 4.47 & 4.79 & 6.09 & 4.88 & 5.28 \\
\midrule
STAG-Net (FaceMesh) & 0.043  & 3.63 & 4.96 & \textbf{2.23} & \underline{3.61} & \underline{3.28} & 4.84 & \underline{3.05} & \underline{3.72} \\
STAG-Net (FAN)      & 0.034  & \textbf{2.86} & 5.00 & \underline{2.68} & \textbf{3.51} & 4.20 & 6.86 & 5.56 & 5.53 \\
\bottomrule
\end{tabular}
}
\end{table*}

\begin{table}[t]
\centering
\caption{Comparison with state of the art methods under Protocol~II,
grouped into RGB image based and landmark based methods, where \textbf{bold}
and \underline{underlined} values indicate the lowest and second lowest
reported numerical values per column across all methods. N/A indicates a value that could not be obtained.}
\label{tab:protocol2}
\resizebox{\columnwidth}{!}{
\begin{tabular}{lccccc}
\toprule
\multirow{2}{*}{\textbf{Method}} &
\multirow{2}{*}{\textbf{Params (M)}} &
\multicolumn{4}{c}{\textbf{BIWI}} \\
\cmidrule(lr){3-6}
& & \textbf{Yaw} & \textbf{Pitch} & \textbf{Roll} & \textbf{MAE} \\
\midrule
\multicolumn{6}{l}{\textit{RGB image based methods}} \\
Martin (2014)\cite{martin}           & N/R      & 3.60 & \underline{2.50} & 2.60 & 2.90 \\
VGG16-RNN (2017) \cite{vgg16} & 138.00  & 3.14 & 3.48 & 2.60 & 3.07 \\
VGG16(2017)\cite{vgg16}             & 138.00  & 3.91 & 4.03 & 3.03 & 3.66 \\
TokenHPE (2023) \cite{tokenhpe}            & 86.42   & 3.01 & \textbf{2.28} & \underline{2.01} & 2.49 \\
6DRepNet (2022) \cite{6drepent} & 39.32 & 2.39 & 2.96 & 2.05 & \underline{2.47} \\
TriNet (2021) \cite{triNet}              & 26.00   & 2.99 & 3.04 & 2.44 & 2.80 \\
FSA-Caps-Fusion (2019) \cite{fsanet}   & 1.20    & 2.89 & 4.29 & 3.60 & 3.60 \\
SSR-Net-MD (2018) \cite{ssr}          & 0.20    & 4.24 & 4.35 & 4.19 & 4.16 \\
LwPosr (2022) \cite{lwposr}              & 0.15    & 3.62 & 4.65 & 3.78 & 4.01 \\
\midrule
\multicolumn{6}{l}{\textit{Landmark based methods}} \\
GADS (2025) \cite{gads}                & 0.02    & 3.31 & 5.00 & 2.94 & 3.75 \\
EVA-GCN (2021) \cite{eva}             & 3.30    & 4.78 & 4.01 & 2.98 & 3.92 \\
\midrule
STAG-Net (FaceMesh) & 0.012 & \underline{2.37} & 3.09 & 2.24 & 2.57 \\
STAG-Net (FAN)      & 0.034   & \textbf{2.31} & 3.00 & \textbf{1.87} & \textbf{2.39} \\
\bottomrule
\end{tabular}
}
\end{table}
The proposed STAG-Net is compared with representative state of the art head pose estimation methods under the two evaluation protocols described in Section~\ref{evaluation_protocols}. Since both RGB based and landmark based approaches have been reported in the literature, comparisons are conducted against methods from both categories to provide a comprehensive evaluation. Furthermore, as STAG-Net can operate with different facial landmark detectors, results are reported using both FaceMesh and FAN landmarks to investigate the influence of the landmark detector on the overall estimation performance. Tables~\ref{tab:protocol1} and~\ref{tab:protocol2} group the compared methods into these two categories, together with the corresponding model size in terms of the number of trainable parameters. For prior methods, the accuracy values and parameter counts are taken directly from the cited publications. Runtime values are measured from the official implementations on common hardware without modifying the models. The preprocessing and pose label conventions in Section~\ref{evaluation_protocols} are applied consistently to both STAG-Net variants. Since the STAG-Net errors are conditional on successful landmark detection, the numerical rankings should be interpreted together with the detector envelopes in Table~\ref{tab:detector_coverage}.

Table~\ref{tab:protocol1} presents the results obtained under Protocol~I on the BIWI and AFLW2000 datasets. On the BIWI dataset, the proposed STAG-Net using FAN landmarks achieves an overall mean absolute error (MAE) of \textbf{3.51$^\circ$}, which is the lowest reported error among the compared methods while requiring only 0.034\,M parameters. Moreover, STAG-Net (FAN) achieves the lowest yaw error among all compared methods, while STAG-Net (FaceMesh) obtains the lowest roll error, demonstrating the effectiveness of the proposed architecture across multiple pose components.

On the AFLW2000 dataset, STAG-Net (FaceMesh) achieves the lowest yaw error (\textbf{3.28$^\circ$}) and the lowest overall MAE (\textbf{3.72$^\circ$}) among the compared landmark based methods, while remaining competitive with the strongest RGB based approaches. Although the FAN based variant achieves superior performance on BIWI, the FaceMesh based variant produces a lower error on AFLW2000, suggesting that the choice of landmark detector can influence performance depending on the characteristics of the target dataset. These results indicate that the proposed architecture is robust across different landmark detectors while maintaining consistently competitive performance.

The results under Protocol~II are summarized in Table~\ref{tab:protocol2}. Both STAG-Net variants outperform existing landmark based approaches despite having substantially fewer trainable parameters. In particular, STAG-Net using FAN achieves the lowest overall MAE of $\mathbf{2.39^\circ}$, together with the lowest yaw and roll errors among all compared methods. STAG-Net using FaceMesh achieves an overall MAE of $\mathbf{2.57^\circ}$ while requiring only 0.012 million trainable parameters. It also obtains the second lowest yaw error of $\mathbf{2.37^\circ}$ among all compared methods and remains more accurate than previously reported landmark based approaches. Together, these results show that STAG-Net remains effective with either landmark detector while providing a favorable balance between accuracy and model size.

An important observation from both evaluation protocols is the computational efficiency of STAG-Net. While several competing methods employ tens or even hundreds of millions of parameters, the proposed models require only \textbf{0.012 to 0.043 million} parameters depending on the landmark detector and configuration (Table~\ref{tab:flops}). Despite this compact design, STAG-Net consistently achieves the lowest reported or highly competitive accuracy across both benchmark protocols. These results demonstrate that the proposed structure aware grouped representation, combined with per group DeepSet encoding and inter group attention, provides an effective balance between estimation accuracy and computational efficiency, making the STAG-Net regressor a promising candidate for TinyML and other resource constrained deployment scenarios.

Because the aggregate errors are computed only for frames with valid landmark detections, the following analysis examines detector coverage and the variation of error across the pose range.
\begin{table}[t]
\centering
\caption{Detection envelope of the two landmark detectors: the range
of ground truth angles (min, max, in degrees) for which each detector
produces valid landmarks, compared with the full annotated range of
each dataset (Table~\ref{tab:pose_ranges}). The regression MAE for each STAG-Net variant is computed only on frames for which its detector returns valid landmarks.}
\label{tab:detector_coverage}
\footnotesize
\setlength{\tabcolsep}{3.5pt}
\resizebox{\columnwidth}{!}{%
\begin{tabular}{llccc}
\toprule
\textbf{Dataset} & \textbf{Angle} & \textbf{Full range} &
\textbf{FaceMesh} & \textbf{FAN} \\
\midrule
\multirow{3}{*}{BIWI}
 & Yaw   & $-74.9,\,78.0$ & $-74.9,\,78.0$ & $-74.9,\,78.0$ \\
 & Pitch & $-63.8,\,84.4$ & $-63.1,\,70.9$ & $-63.8,\,82.4$ \\
 & Roll  & $-64.7,\,45.0$ & $-48.7,\,43.2$ & $-56.4,\,45.0$ \\
\midrule
\multirow{3}{*}{300W-LP}
 & Yaw   & $-90.0,\,90.0$ & $-90.0,\,89.9$ & $-90.0,\,90.0$ \\
 & Pitch & $-64.6,\,61.9$ & $-59.4,\,58.3$ & $-63.9,\,61.1$ \\
 & Roll  & $-73.4,\,53.2$ & $-68.5,\,51.9$ & $-73.4,\,53.2$ \\
\midrule
\multirow{3}{*}{AFLW2000}
 & Yaw   & $-83.7,\,86.6$ & $-80.6,\,83.1$ & $-83.7,\,86.6$ \\
 & Pitch & $-83.0,\,88.3$ & $-64.6,\,62.5$ & $-83.0,\,88.3$ \\
 & Roll  & $-87.1,\,83.7$ & $-86.2,\,80.3$ & $-86.5,\,83.7$ \\
\bottomrule
\end{tabular}%
}
\end{table}

\begin{figure*}[t]
\centering
\includegraphics[width=\textwidth]{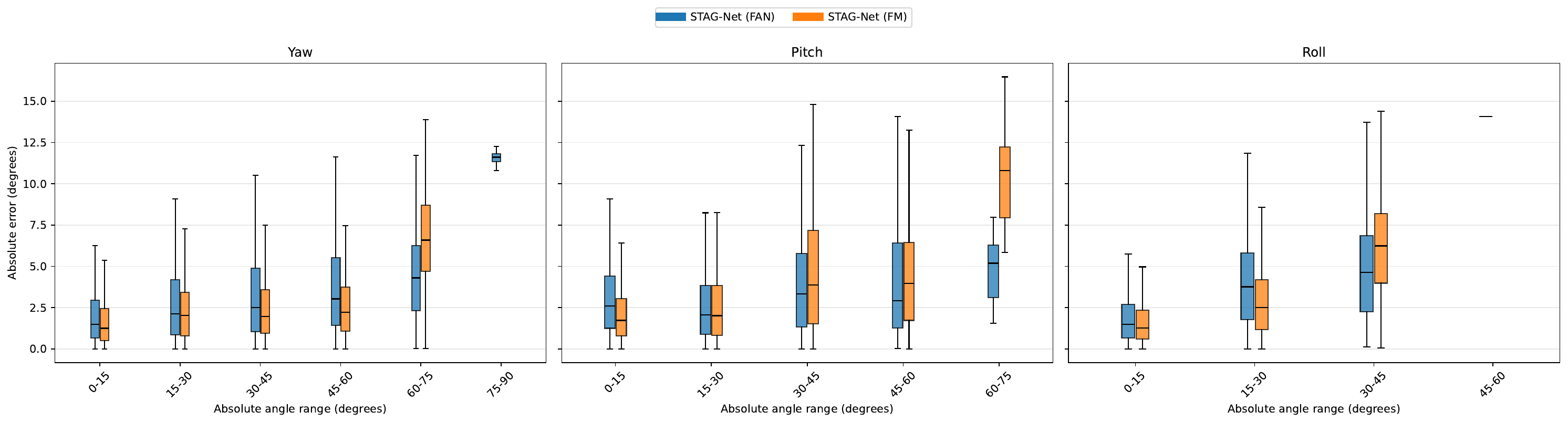}
\caption{Distribution of absolute errors of the two STAG-Net variants
on BIWI (Protocol~I), binned by ground truth angle magnitude. A
missing box indicates that the corresponding landmark detector
produced too few valid detections in that range to form an error
distribution, consistent with the detector envelopes reported in
Table~\ref{tab:detector_coverage}. Per bin MAE curves are provided in
Appendix~\ref{app:error_curves}.}
\label{fig:error_boxplots}
\end{figure*}

\paragraph{Per Angle Error Analysis}
Fig.~\ref{fig:error_boxplots} shows how the absolute error changes
with ground truth angle magnitude on BIWI, while
Table~\ref{tab:detector_coverage} reports the angle ranges for which
each detector returns valid landmarks. Both variants are most
accurate near frontal poses and their errors increase as the pose
magnitude grows. This pattern is consistent with the concentration of
300W-LP samples around frontal poses, as shown in
Fig.~\ref{fig:angle_histograms}, and with the increasing
self occlusion of facial landmarks at large rotations.

The two variants differ mainly because their detectors have different
operating ranges. FaceMesh is generally more accurate over the
moderate pose ranges that contain most of the test samples, while FAN
covers more of the extreme pose range. This difference is visible in
the outermost bins of Fig.~\ref{fig:error_boxplots}. In pitch, both
detectors cover every bin, since the FaceMesh envelope extends to
$70.9^\circ$. In yaw and roll, however, the FaceMesh envelope ends at
$78.0^\circ$ and $48.7^\circ$ respectively, which lies just inside the
lower edge of the largest bin in each case; as a result, FaceMesh
yields too few detections there to form an error distribution, and
only the FAN variant produces predictions across the full annotated
range. FAN errors also increase near the limits of its own coverage.
The combined system therefore presents a detector level tradeoff.
FaceMesh offers higher throughput and strong performance in near
frontal and moderate poses, while FAN provides broader angular coverage at a much higher computational cost. This finding explains
why detector choice affects both the aggregate errors in
Tables~\ref{tab:protocol1} and~\ref{tab:protocol2} and the practical
operating range of the deployed pipeline.

The comparison with existing methods demonstrates that STAG-Net achieves strong estimation accuracy while retaining a compact model size across both evaluation protocols. However, these results alone do not explain how the individual design choices contribute to the final performance. Therefore, the following subsection examines the influence of landmark selection, connectivity graph construction, architectural configuration, and learning rate scheduling to identify the design choices that support the final STAG-Net configuration.

\subsection{Design Analysis}
\label{sec:design}

The following experiments examine STAG-Net using FaceMesh under Protocol~II. Unless otherwise stated, only the component under investigation is changed while the remaining clustering, architectural, and training settings are kept fixed.

\paragraph{Effect of the Connectivity Graph Neighborhood}

The connectivity graph used during agglomerative clustering determines which facial landmarks are allowed to be grouped together. To examine its influence, the FaceMesh connectivity graph is constructed using two, four, and six nearest neighbors while keeping the clustering procedure, movement based landmark selection, model architecture, and training settings unchanged. The number of clusters is not manually fixed for this comparison. Instead, it is selected independently using silhouette analysis. As shown in Table~\ref{tab:abl_neighbors}, all three neighborhood settings produce five clusters.

\begin{table}[!t]
\caption{Comparison of connectivity graph neighborhood settings for STAG-Net using FaceMesh under Protocol~II. The number of clusters is selected using silhouette analysis.}
\label{tab:abl_neighbors}
\centering
\setlength{\tabcolsep}{4pt}
\renewcommand{\arraystretch}{1.1}
\begin{tabular}{cccccc}
\toprule
\textbf{Neighbors} &
\textbf{Clusters} &
\textbf{Yaw} &
\textbf{Pitch} &
\textbf{Roll} &
\textbf{MAE} \\
\midrule
2 & 5 & 2.517 & 3.218 & 2.367 & 2.701 \\
4 & 5 & 2.440 & 3.235 & 2.296 & 2.657 \\
6 & 5 & \textbf{2.373} & \textbf{3.091} & \textbf{2.239} & \textbf{2.568} \\
\bottomrule
\end{tabular}
\end{table}

\begin{figure*}[!t]
    \centering

    \begin{subfigure}[t]{0.32\textwidth}
        \centering
        \includegraphics[width=\linewidth]{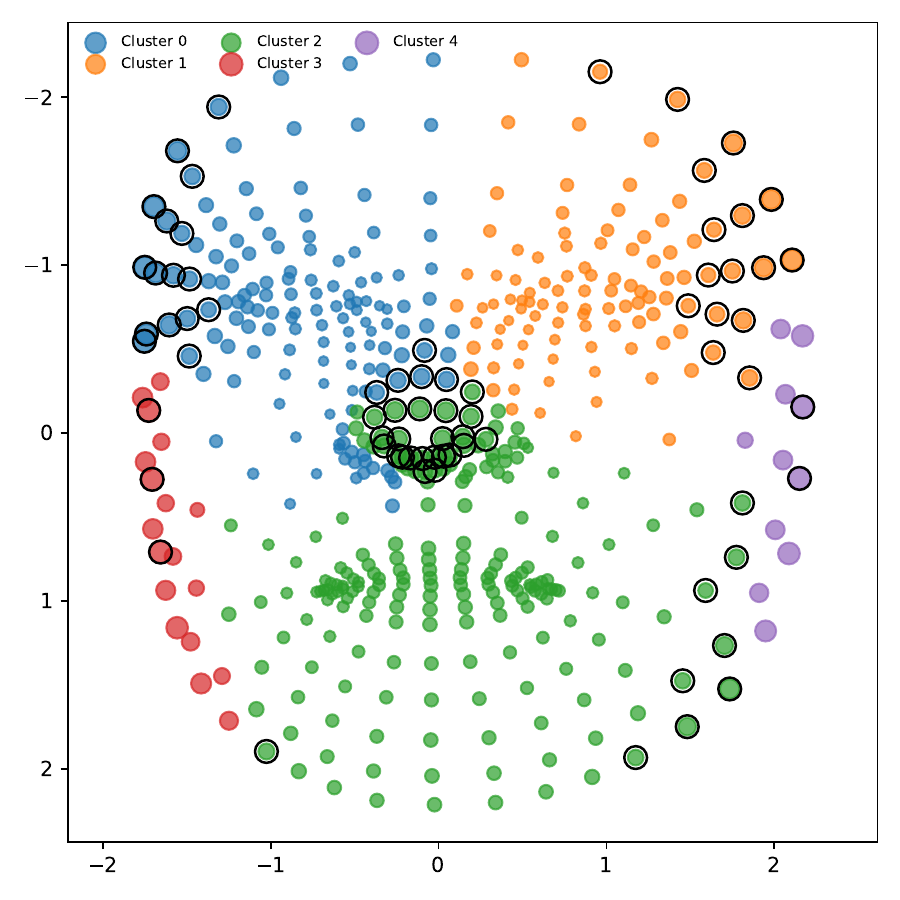}
        \caption{Two neighbors}
        \label{fig:neighbors_2}
    \end{subfigure}
    \hfill
    \begin{subfigure}[t]{0.32\textwidth}
        \centering
        \includegraphics[width=\linewidth]{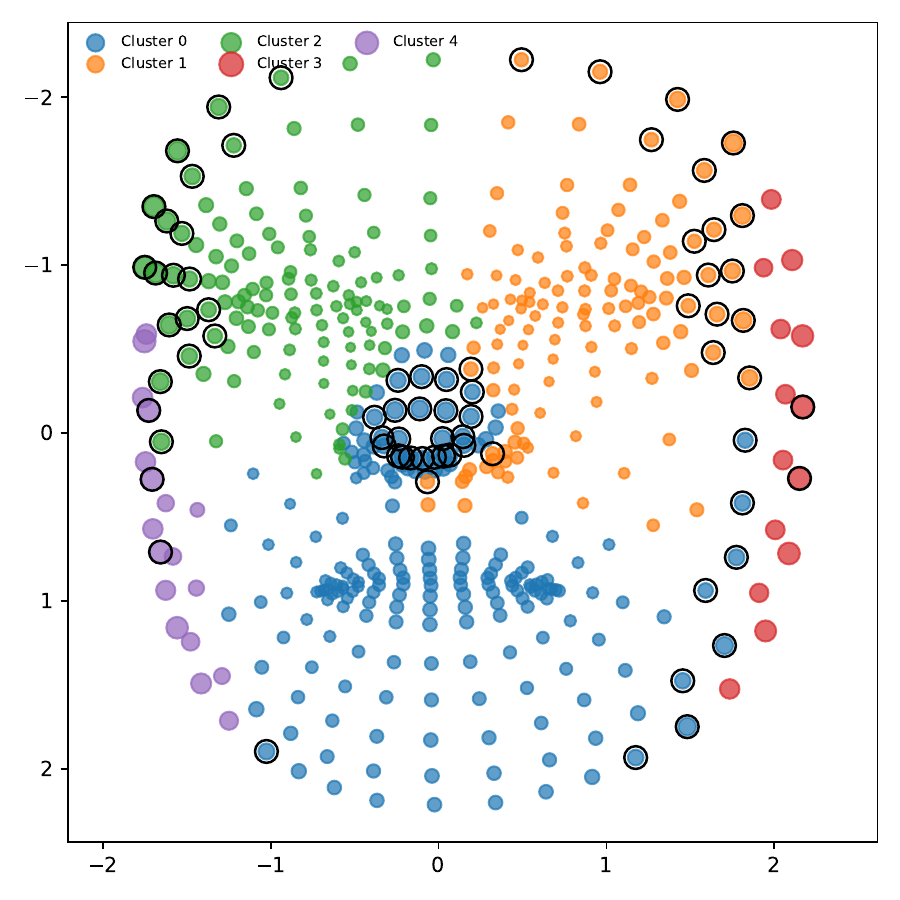}
        \caption{Four neighbors}
        \label{fig:neighbors_4}
    \end{subfigure}
    \hfill
    \begin{subfigure}[t]{0.32\textwidth}
        \centering
        \includegraphics[width=\linewidth]{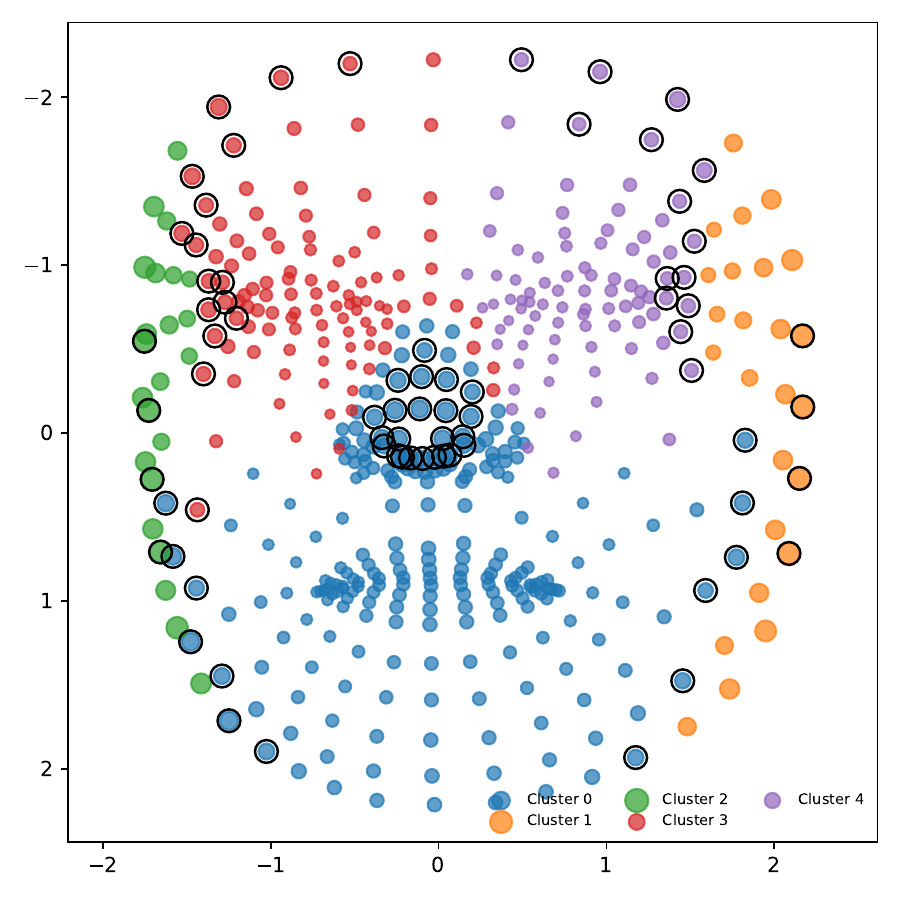}
        \caption{Six neighbors}
        \label{fig:neighbors_6}
    \end{subfigure}

    \caption{Automatically generated FaceMesh landmark clusters obtained using connectivity graphs with different neighborhood sizes. Black outlines indicate landmarks whose movement scores are at or above the 85th percentile. Silhouette analysis produces five clusters in all three settings, while the neighborhood size changes their spatial composition.}
    \label{fig:connectivity_neighbors}
\end{figure*}

The visualizations in Fig.~\ref{fig:connectivity_neighbors} show that changing the neighborhood size alters the spatial composition of the automatically generated clusters, although silhouette analysis selects five clusters in every case. The four and six neighbor settings produce more spatially continuous groupings than the two neighbor setting. This observation is consistent with the quantitative results in Table~\ref{tab:abl_neighbors}. Increasing the neighborhood size from two to four reduces the overall MAE from $2.701$ to $2.657$, while six neighbors further reduce it to $2.568$. The six neighbor setting also achieves the lowest yaw, pitch, and roll errors among the evaluated configurations. Therefore, six neighbors are used to construct the connectivity graph in the final STAG-Net configuration.

Changing the graph neighborhood does not alter the trainable parameter count because the model architecture remains unchanged and all three settings produce five clusters. The neighborhood size only affects the connectivity constraints used during clustering and consequently changes the cluster membership and the landmarks retained from each cluster.

\paragraph{Effect of the Landmark Selection Percentage}

\begin{table}[!t]
\caption{Effect of movement based landmark selection for STAG-Net using FaceMesh under Protocol~II using six connectivity graph neighbors and the resulting five clusters. Boldface identifies the configuration selected according to the lowest overall MAE.}
\label{tab:abl_landmark_selection}
\centering
\begin{tabular}{ccccc}
\toprule
\textbf{Retained} &
\textbf{Yaw} &
\textbf{Pitch} &
\textbf{Roll} &
\textbf{MAE} \\
\midrule
25\% & 2.408 & 3.162 & 2.184 & 2.584 \\
20\% & 2.334 & 3.242 & 2.276 & 2.617 \\
\textbf{15\%} & \textbf{2.373} & \textbf{3.091} &
\textbf{2.239} & \textbf{2.568} \\
10\% & 2.428 & 3.180 & 2.329 & 2.646 \\
5\%  & 2.572 & 2.930 & 2.421 & 2.641 \\
\bottomrule
\end{tabular}%
\end{table}

Table~\ref{tab:abl_landmark_selection} examines the effect of movement based landmark selection for STAG-Net using FaceMesh under Protocol~II. Retaining the highest 15\% of movement ranked landmarks produces the lowest overall MAE of $\mathbf{2.568}$ and is therefore used in the main experiments, as indicated by the bolded row. Retaining 25\% produces the lowest roll error and the second lowest overall MAE of $2.584$, while retaining 20\% produces the lowest yaw error. Retaining only 5\% produces the lowest pitch error, but its yaw and roll errors are higher. These results show that the optimal retained percentage varies across the individual pose angles. However, retaining 15\% provides the lowest average error across yaw, pitch, and roll.

\paragraph{Effect of Architectural Hyperparameters}
\begin{table}[t]
\centering
\caption{Effect of architectural hyperparameters (STAG-Net (FaceMesh),
Protocol~II). Defaults marked $\dagger$: dropout $0.01$, $n_h=2$,
$h_e=32$, one encoder and one decoder layer.}
\label{tab:abl_arch}
\small
\resizebox{\columnwidth}{!}{%
\begin{tabular}{llcccc}
\toprule
 & Setting & \textbf{Yaw} & \textbf{Pitch} & \textbf{Roll} & \textbf{MAE} \\
\midrule
\multirow{3}{*}{Dropout}
 & 0               & 2.60 & 3.05 & 2.31 & 2.65 \\
 & 0.01$^\dagger$  & 2.37 & 3.09 & 2.24 & \textbf{2.57} \\
 & 0.05             & 2.49 & 3.26 & 2.42 & 2.72 \\
\midrule
\multirow{3}{*}{Heads $n_h$}
 & 2$^\dagger$ & 2.37 & 3.09 & 2.24 & \textbf{2.57} \\
 & 4           & 2.44 & 3.26 & 2.29 & 2.67 \\
 & 6           & 2.40 & 3.26 & 2.34 & 2.66 \\
\midrule
\multirow{2}{*}{Width $h_e$}
 & 32$^\dagger$ & 2.37 & 3.09 & 2.24 & \textbf{2.57} \\
 & 64           & 2.46 & 3.22 & 2.25 & 2.64 \\
\midrule
\multirow{4}{*}{Encoder and decoder layers}
 & 1 and 1$^\dagger$ & 2.37 & 3.09 & 2.24 & \textbf{2.57} \\
 & 1 and 2           & 2.57 & 3.04 & 2.37 & 2.66 \\
 & 1 and 3           & 2.48 & 3.05 & 2.34 & 2.62 \\
 & 2 and 2           & 2.56 & 3.09 & 2.41 & 2.69 \\
\bottomrule
\end{tabular}%
}
\end{table}
Table~\ref{tab:abl_arch} examines the sensitivity of STAG-Net using FaceMesh under Protocol~II to its main architectural choices. The model remains robust across all evaluated settings, with the overall MAE ranging from \(2.57^\circ\) to \(2.72^\circ\). The selected configuration, consisting of a dropout rate of 0.01, two attention heads, an encoder width of 32, and one encoder and one decoder layer, achieves the lowest overall MAE of \(2.57^\circ\). Increasing the number of attention heads, widening the encoder, or deepening the encoder and decoder stack does not improve the overall accuracy. These findings support the use of the smallest evaluated architectural configuration.

\paragraph{Effect of the Learning Rate Scheduling Strategy}
To examine the effect of the optimization schedule, four learning rate strategies are compared using STAG-Net with FaceMesh under Protocol~II. All remaining training and architectural settings are kept unchanged. CosineAnnealingLR is applied over the complete 150-epoch training period with \(T_{\max}=150\), while MultiStepLR reduces the learning rate at epochs 70, 90, and 120. The constant learning rate setting retains the initial learning rate throughout training. CosineAnnealingWarmRestarts uses an initial restart period of \(T_0=10\) and a restart period multiplier of \(T_{\mathrm{mult}}=2\).

\begin{table}[!t]
\caption{Comparison of learning rate scheduling strategies for STAG-Net using FaceMesh under Protocol~II.}
\label{tab:scheduler_comparison}
\centering
\small
\resizebox{\columnwidth}{!}{%
\begin{tabular}{lcccc}
\toprule
\textbf{Scheduler} &
\textbf{Yaw} &
\textbf{Pitch} &
\textbf{Roll} &
\textbf{MAE} \\
\midrule
CosineAnnealingWarmRestarts
& 2.373 & 3.091 & 2.239 & \textbf{2.568} \\

CosineAnnealingLR
& 2.456 & \textbf{3.079} & \textbf{2.192} & 2.576 \\

MultiStepLR
& 3.576 & 3.538 & 2.351 & 3.155 \\

Constant learning rate
& \textbf{2.365} & 3.100 & 2.307 & 2.591 \\
\bottomrule
\end{tabular}
}
\end{table}

Table~\ref{tab:scheduler_comparison} shows that CosineAnnealingWarmRestarts achieves the lowest overall MAE of $\mathbf{2.568}$ and is therefore used in the main experiments. The constant learning rate produces the lowest yaw error of $\mathbf{2.365}$, while CosineAnnealingLR produces the lowest pitch and roll errors of $\mathbf{3.079}$ and $\mathbf{2.192}$, respectively. However, CosineAnnealingWarmRestarts provides the most balanced performance across the three pose angles. MultiStepLR performs considerably worse, with an overall MAE of $3.155$. The relatively small differences among CosineAnnealingWarmRestarts, CosineAnnealingLR, and the constant learning rate indicate that STAG-Net remains stable across these scheduling strategies, although the warm restart schedule provides the strongest overall result.

Overall, the design analysis shows that the proposed movement aware landmark selection, connectivity based clustering, compact architecture, and learning rate scheduling strategy each contribute to the final performance of STAG-Net. The results indicate that retaining only the most informative landmarks is sufficient for accurate estimation and that increasing architectural complexity does not consistently improve performance. The clustering and architectural configurations vary according to the landmark detector and evaluation protocol, while all variants retain the same lightweight design principles. The complete detector and protocol specific STAG-Net configurations are summarized in Appendix~\ref{app:stagnet_configurations}. Having established the final design choices, the following subsection evaluates the computational performance of the resulting STAG-Net models.

\subsection{Computational Performance Analysis}
\label{performance_results}

\paragraph{Model Complexity}
Table~\ref{tab:flops} compares the trainable parameters and inference
FLOPs of STAG-Net with representative image based methods. Parameter
counts for prior methods follow their original publications, while
FLOPs are measured from the released implementations where available.
For STAG-Net, the reported FLOPs cover only the pose regression model.
Across the four configurations, the regressor requires between
$2.4\times10^{4}$ and $8.4\times10^{4}$ FLOPs per inference. This is
approximately three orders of magnitude below FSA-Net and five to
six orders of magnitude below HopeNet, 6DRepNet, and TokenHPE. The largest STAG-Net
configuration requires about $170$ KB for 32 bit parameters and about
$43$ KB for 8 bit parameters, before accounting for implementation
and activation memory.

These figures describe the compactness of the STAG-Net regressor, not
the cost of landmark extraction. The combined pipeline cost depends
on the detector. FaceMesh supports real time CPU
inference in our measurements, while FAN is considerably more expensive. In applications
where landmarks are already produced for face tracking or expression
analysis, STAG-Net adds only a small incremental cost. A complete
microcontroller deployment, including landmark extraction and measured
activation memory, remains outside the scope of the present study.

\begin{figure*}[t]
\centering
\includegraphics[width=\textwidth]{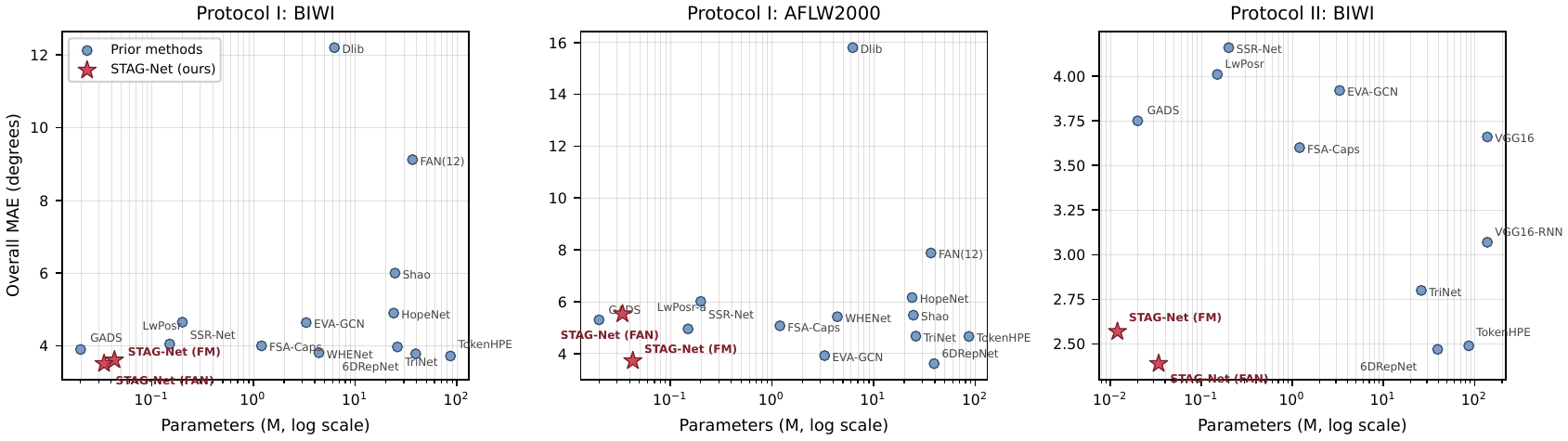}
\caption{Accuracy and model size trade-off: overall MAE versus
trainable parameters (log scale) for all compared methods with
published parameter counts; FM denotes the FaceMesh variant. STAG-Net
variants are marked with stars; methods toward the lower left are
jointly smaller and more accurate.}

\label{fig:pareto}
\end{figure*}

\paragraph{Pareto Analysis of Accuracy and Model Size}
Fig.~\ref{fig:pareto} combines the accuracy and parameter results
from Tables~\ref{tab:protocol1} and~\ref{tab:protocol2}. Under
Protocol~II, the Pareto frontier contains only the two STAG-Net
variants. The FaceMesh variant provides the smallest model and the FAN
variant provides the lowest error. Under Protocol~I on BIWI, GADS and STAG-Net using FAN form the frontier. The FAN configuration is both smaller and more accurate than the FaceMesh configuration in this setting. On AFLW2000, 6DRepNet is $0.11^\circ$ more
accurate than STAG-Net using FaceMesh, but it uses roughly 900 times
more parameters. The Pareto analysis therefore shows that STAG-Net
achieves compactness without simply accepting a large loss in
accuracy.

\begin{table}[t]
\centering
\caption{Model complexity comparison. STAG-Net FLOPs are reported per
variant (Protocol~I / Protocol~II) and cover the pose regression model
only, while RGB based baselines process raw images directly.}
\label{tab:flops}
\small
\begin{tabularx}{\columnwidth}{@{}>{\raggedright\arraybackslash}Xrr@{}}
\toprule
\textbf{Method} & \textbf{Params (M)} & \textbf{FLOPs} \\
\midrule
TokenHPE            & 86.42  & $33.97$\,G \\
6DRepNet            & 39.32  & $9.86$\,G \\
HopeNet             & 23.90  & $4.11$\,G  \\
FSA-Net             & 1.20   & $91.58$\,M \\
\midrule
STAG-Net (FaceMesh), Protocol~I   & 0.043 & $83.6$\,K \\
STAG-Net (FAN), Protocol~I  & 0.034 & $70.7$\,K \\
STAG-Net (FaceMesh), Protocol~II  & 0.012 & $24.1$\,K \\
STAG-Net (FAN), Protocol~II & 0.034 & $70.5$\,K \\
\bottomrule
\end{tabularx}
\end{table}

\paragraph{Inference Performance}
\begin{table}[t]
\centering
\caption{Inference throughput (frames per second) measured on the
Google Colab CPU runtime and an NVIDIA Tesla T4 GPU. For STAG-Net,
throughput is reported for the complete pipeline from a cropped face
image, covering image conversion, landmark extraction, normalization,
and regression, and is measured using the largest configuration of
each variant (Protocol~I), so the reported values represent the worst
case across configurations. RGB based baselines are measured from the
same cropped face input as in the original works.}
\label{tab:inference_speed}
\small
\begin{tabularx}{\columnwidth}{@{}>{\raggedright\arraybackslash}Xrr@{}}
\toprule
\textbf{Method} & \textbf{CPU (FPS)} & \textbf{GPU (FPS)} \\
\midrule
HopeNet          & 4.4    & 147.3 \\
6DRepNet         & 6.9   & 62.0  \\
FSA-Net          & 114.0 & 164.5 \\
TokenHPE         & 6.6    & 69.9  \\
\midrule
STAG-Net (FaceMesh)    & 111.0 & 322.9 \\
STAG-Net (FAN)   & 1.6   & 17.7  \\
\bottomrule
\end{tabularx}
\end{table}

Table~\ref{tab:inference_speed} reports throughput on the Colab CPU and Tesla T4 GPU described in Section~\ref{implementation_details}. For STAG-Net, the measurement starts from a cropped face image and includes image conversion, landmark extraction, normalization, landmark selection, and pose regression. The measurements use the largest configuration of each variant, so the reported throughput is conservative. The reported frame rates therefore represent the landmark extraction and regression pipeline rather than the regressor alone.

The FaceMesh pipeline reaches 322.9 FPS on the GPU and 111.0 FPS on the CPU. Its GPU throughput is almost twice that of the fastest measured image based baseline, while its CPU throughput is close to the 114.0 FPS measured for FSA-Net on the same runtime. Notably, the CPU throughput of the FaceMesh pipeline also exceeds the GPU throughput of 6DRepNet, which is the closest competitor in accuracy under Protocol~II and reaches only 6.9 FPS on the CPU. The FAN pipeline is much slower, reaching 17.7 FPS on the GPU and 1.6 FPS on the CPU. This difference is mainly caused by landmark extraction because the STAG-Net regression stage requires fewer than $8.4\times10^{4}$ FLOPs. The two variants therefore provide a practical choice between throughput and angular coverage. FaceMesh is the stronger option for latency constrained applications, while FAN is useful when broader extreme pose coverage is more important.

\paragraph{Training Efficiency}
Table~\ref{tab:train_time} reports the per epoch training time on the
hardware described in Section~\ref{implementation_details}. Because STAG-Net operates on
previously extracted landmark coordinates rather than raw images, a full
epoch over the 300W-LP training set ($\approx$122{,}450 samples)
completes in under 11 seconds on the T4 GPU and under 29 seconds on
the CPU; training for all 150 epochs therefore finishes in
between 24 and 28 minutes on the GPU, or about 70 minutes on CPU alone. Since
the baselines are trained on the much smaller BIWI split
($\approx$11{,}000 frames), raw per epoch times are not directly
comparable; normalizing by dataset size, STAG-Net processes a training
sample roughly $7$ to $8\times$ faster than FSA-Net and two orders of
magnitude faster than HopeNet, 6DRepNet, and TokenHPE on the same GPU. This low
training cost makes hyperparameter exploration and retraining for new
deployment conditions practical without dedicated hardware.

\begin{table}[t]
\centering
\caption{Training time per epoch. The last column normalizes the GPU time by dataset size for direct comparison.}
\label{tab:train_time}
\small
\begin{tabularx}{\columnwidth}{@{}>{\raggedright\arraybackslash}Xrrr@{}}
\toprule
\textbf{Method} & \textbf{CPU (s)} & \textbf{GPU (s)} & \shortstack{\textbf{GPU}\\\textbf{(ms/sample)}} \\
\midrule
HopeNet & 8051.9 & 127.5 & 11.6 \\
6DRepNet & 5232.2 & 253.2 & 23.0 \\
TokenHPE & 6365.5 & 410.3 & 37.4 \\
FSA-Net &  203.1 &   6.7 &  0.61 \\
\midrule
STAG-Net (FaceMesh) &   25.9 &   9.5 &  0.077 \\
STAG-Net (FAN) &   28.0 &  10.9 &  0.089 \\
\bottomrule
\end{tabularx}
\end{table}

\paragraph{Summary of Findings}
The experiments support four conclusions. First, grouped set
processing is effective for landmark based head pose estimation.
STAG-Net achieves the lowest reported MAE among the compared landmark based methods across all three evaluation settings and the lowest reported BIWI MAE among all compared methods under both protocols on frames with valid landmark detections. Second,
the regression model is compact in terms of parameters and FLOPs, and
the FaceMesh configuration provides real time throughput for landmark extraction and regression on the Google Colab CPU runtime. Third, the design analysis shows that
performance is stable across the tested architectural settings. It
also confirms that a compact landmark subset selected using movement is
sufficient and that connectivity graph construction influences the
quality of the groups. Fourth, the main variation between the two
system configurations comes from the landmark detector. Detector
coverage limits estimation at extreme poses and detector cost
dominates the latency of the combined pipeline.

\section{Conclusion}
\label{sec:conclusion}

This work introduced STAG-Net, a compact structure aware architecture
for landmark based head pose estimation. The method groups facial
landmarks into automatically discovered spatial regions, retains the
most pose sensitive landmarks in each region, encodes every group as
an unordered set, and models relationships among groups through
multihead self attention. A compact geometry descriptor complements
the learned group representation before the final yaw, pitch, and
roll prediction.

The results follow the objectives established in the introduction.
Across the three reported evaluation settings, STAG-Net achieves the lowest reported MAE among the landmark based methods included in the comparison. On frames with valid landmark detections, it also obtains the lowest reported overall BIWI MAE among all compared methods under both protocols, with $3.51^\circ$ under cross dataset evaluation and $2.39^\circ$ under subject independent evaluation. Depending on the configuration, the pose regressor uses
between $0.012$ and $0.043$ million parameters and requires fewer than
$8.4\times10^{4}$ FLOPs per inference. The Pareto analysis further
shows that the two STAG-Net variants alone form the Protocol~II
frontier. The FaceMesh configuration provides real time throughput for landmark extraction and regression on the Google Colab CPU runtime, while FAN provides wider
coverage at extreme poses. The design analysis supports the movement
aware selection strategy and shows that the compact architecture does
not depend on a narrowly tuned hyperparameter setting. We also
identified and removed 36 physically implausible AFLW2000 annotations
and reported the cleaned pose ranges used in the evaluation.

Two limitations remain. First, the complete system inherits the
coverage and computational cost of its landmark detector. FaceMesh is
fast but does not cover some extreme poses, while FAN covers a wider
range but is not suitable for real time CPU operation. Second, the
TinyML assessment is based on parameter counts, regression FLOPs, and
Google Colab CPU measurements rather than a complete microcontroller
implementation. Future work should therefore evaluate quantized
STAG-Net models on microcontroller class hardware and report measured
latency, flash use, activation memory, and energy. A lightweight on
device landmark detector is also required before the entire pipeline
can be considered fully embedded. Further work can extend the grouped
set representation to continuous rotation outputs and to related
landmark based tasks such as gaze and expression estimation.

\section*{Acknowledgment}

This study is funded by the Accelerating Higher Education Expansion
and Development (AHEAD) Operation of the Ministry of
Higher Education of Sri Lanka, funded by the World Bank
(\url{https://ahead.lk/result-area-3/}) and a research grant funded by the Sri Lanka Institute of Information Technology, Sri Lanka (Grant No.~PVC(R\&I)/RG/2025/12). .

The authors used Anthropic's Claude to assist with improving grammar and clarity, generating the model architecture diagram, and checking numerical consistency across tables. All experiments, data, technical content, analyses, results,
interpretations, and conclusions were developed and verified by the authors, who take full responsibility for the content of this article.

\bibliographystyle{IEEEtran}
\bibliography{references}

j

\newpage

\appendices
\section{Per Bin Error Curves}
\label{app:error_curves}
\begin{figure}[ht]\centering
\includegraphics[width=0.8\linewidth]{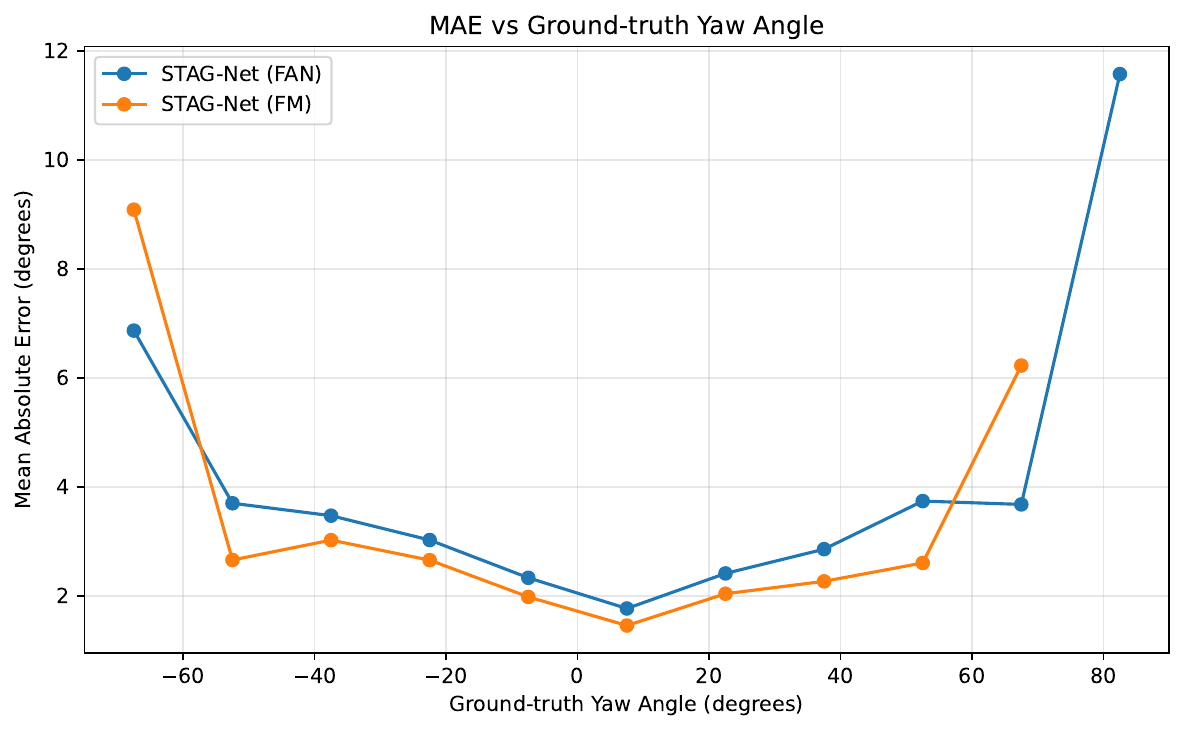}
\caption{MAE versus ground truth yaw angle.}\label{fig:mae_yaw}
\end{figure}

\begin{figure}[ht]\centering
\includegraphics[width=0.8\linewidth]{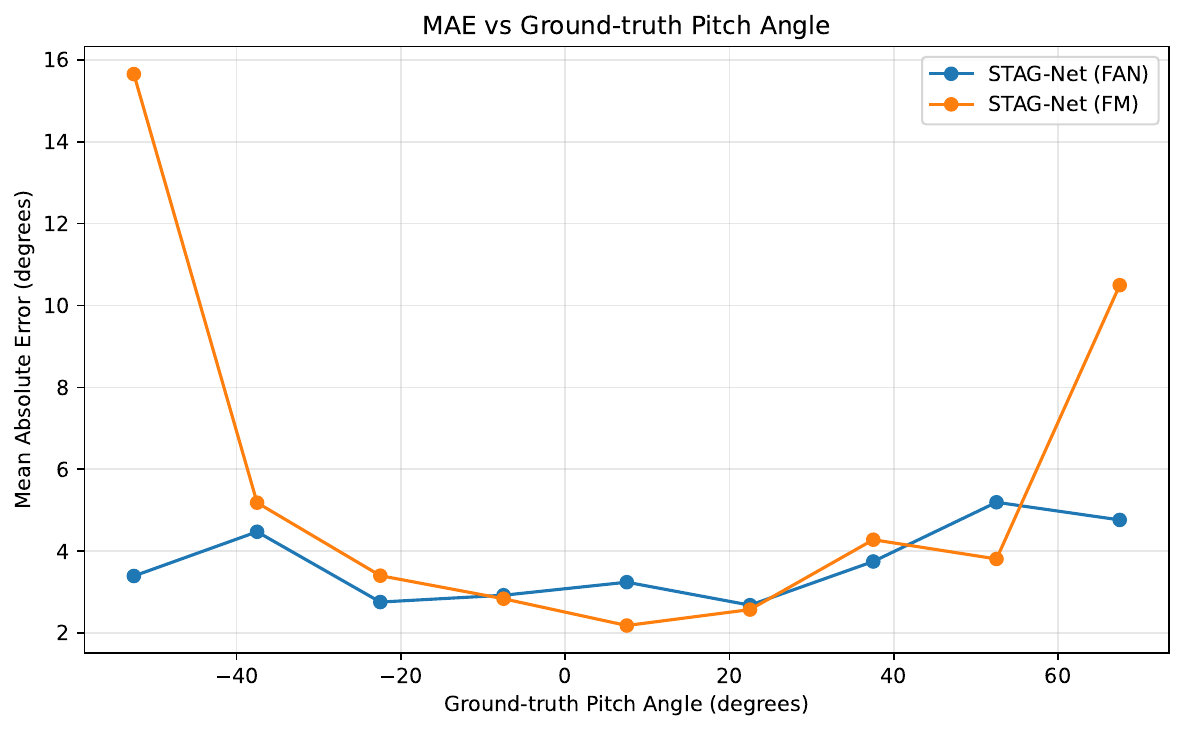}
\caption{MAE versus ground truth pitch angle.}\label{fig:mae_pitch}
\end{figure}

\begin{figure}[ht]\centering
\includegraphics[width=0.8\linewidth]{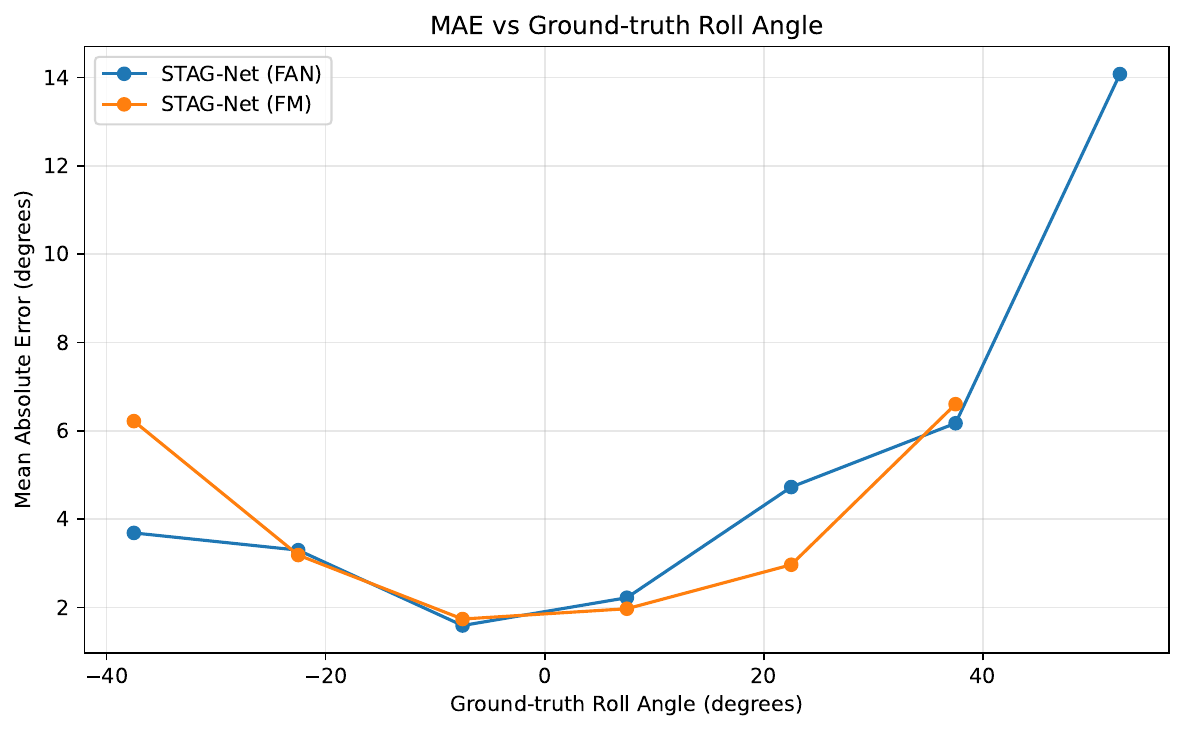}
\caption{MAE versus ground truth roll angle.}\label{fig:mae_roll}
\end{figure}

Figs.~\ref{fig:mae_yaw},~\ref{fig:mae_pitch}, and~\ref{fig:mae_roll} complement the box plots
in Fig.~\ref{fig:error_boxplots} by showing the mean absolute error
of both STAG-Net variants as a function of the ground truth yaw,
pitch, and roll angle on BIWI. Gaps in a curve correspond to angle
ranges in which the respective landmark detector produced too few valid detections to compute a reliable mean.


\section{STAG-Net Configuration Details}
\label{app:stagnet_configurations}

Table~\ref{tab:arch-config} summarizes the detector and protocol specific configurations used in the final experiments. Each cluster is processed using an independent DeepSet encoder consisting of input batch normalization, a linear projection to \(h_e\) units, batch normalization, ReLU activation, and dropout. Mean and max pooling are then applied to obtain a fixed length representation for each cluster. The decoder projects the pooled representation to \(F\) units, while each angle specific regression head maps the trunk output from \(D\) to 32 and then to a single pose estimate.

For a fixed encoder width, the use of an independent DeepSet encoder for each cluster causes the encoder parameter count to increase with the number of clusters \(K\). However, the total parameter count also depends on the encoder width, token dimension, and trunk output dimension. The number of attention heads partitions the token representation and therefore does not change the parameter count when the token dimension remains fixed.

\begin{table}[!t]
\centering
\caption{Detector and protocol specific STAG-Net architecture configurations.}
\label{tab:arch-config}
\footnotesize
\setlength{\tabcolsep}{3.5pt}
\renewcommand{\arraystretch}{1.1}

\begin{tabularx}{\columnwidth}{
@{}
>{\raggedright\arraybackslash}X
cccc
@{}}
\toprule
&
\multicolumn{2}{c}{\textbf{Protocol I}} &
\multicolumn{2}{c}{\textbf{Protocol II}} \\
\cmidrule(lr){2-3}
\cmidrule(lr){4-5}

\textbf{Configuration} &
\textbf{FaceMesh} &
\textbf{FAN} &
\textbf{FaceMesh} &
\textbf{FAN} \\
\midrule

Encoder width, \(h_e\)
& 64 & 64 & 32 & 64 \\

Token dimension, \(F\)
& 32 & 32 & 16 & 32 \\

Attention heads, \(n_h\)
& 4 & 4 & 2 & 2 \\

Clusters, \(K\)
& 7 & 5 & 5 & 5 \\

Trunk output, \(D\)
& 36 & 36 & 32 & 36 \\

Connectivity graph neighbors
& 6 & 4 & 6 & 4 \\

Retained landmarks, \(N_s\) (\%)
& 15 & 40 & 15 & 40 \\

Trainable parameters
& 42{,}557
& 33{,}521
& 12{,}209
& 33{,}521 \\

\bottomrule
\end{tabularx}
\end{table}

\EOD

\end{document}